\documentclass{article} % For LaTeX2e

\usepackage{iclr2027_conference}
\usepackage{times}

\usepackage{newtxtt} % \texttt
\usepackage[scaled=1]{biolinum} % \textsf
\usepackage{graphicx}
\usepackage{booktabs} % for professional tables
\usepackage[font={footnotesize}]{caption}
\usepackage{subcaption} % For subfigures
\usepackage{float} 
\usepackage[utf8]{inputenc} % allow utf-8 input
\usepackage[T1]{fontenc}    % use 8-bit T1 fonts
\usepackage[colorlinks=true, citecolor=Violet, linkcolor=red, urlcolor=magenta]{hyperref}       % hyperlinks
\usepackage{url}            % simple URL typesetting
\usepackage{booktabs}   

\usepackage{amsfonts}       % blackboard math symbols
\usepackage{nicefrac}       % compact symbols for 1/2, etc.
\usepackage{microtype}      % microtypography
\usepackage{lipsum}
\usepackage{fancyhdr}       % header
\usepackage{wrapfig}
\usepackage{colortbl}
\usepackage[dvipsnames]{xcolor}
\usepackage{graphicx}    % graphics
\graphicspath{{media/}}     % organize your images and other figures under media/ folder
\usepackage[most]{tcolorbox}
\usepackage{alltt}
\usepackage{enumitem}
\usepackage{bm}
\usepackage{bbm}
\usepackage{tikz}
\usetikzlibrary{arrows.meta}

\usepackage{amsmath}
\usepackage{amssymb}
\usepackage{mathtools}
\usepackage{amsthm}
\usepackage{amsfonts} 
\usepackage{thmtools, thm-restate} % restate theorem

\usepackage[capitalize,noabbrev]{cleveref}

\usepackage[vlined,ruled,linesnumbered]{algorithm2e}
\usepackage{adjustbox}
\usepackage{multirow,mathtools}

\usepackage{etoc}

\usepackage{threeparttable}
\usepackage{arydshln} % hdashline

\theoremstyle{plain}
\newtheorem{theorem}{Theorem}[section]
\newtheorem{proposition}[theorem]{Proposition}

\theoremstyle{definition}

\theoremstyle{remark}

\newtcbox{\hlprimarytab}{on line, box align=base, colback=BlueGreen!20,colframe=blue,size=fbox,arc=3pt, before upper=\strut, top=-2.5pt, bottom=-4.5pt, left=-2pt, right=-2pt, boxrule=0pt}
\newtcbox{\hlsecondarytab}{on line, box align=base, colback=WildStrawberry!10,colframe=orange,size=fbox,arc=3pt, before upper=\strut, top=-2.5pt, bottom=-4.5pt, left=-2pt, right=-2pt, boxrule=0pt}

\newcommand{\refs}[2]{\hyperref[#1]{\ref*{#1}#2}}

\newtcolorbox[auto counter, number within=section]{mydefinition}[2][]{%
  enhanced,
  breakable,
  colback=green!1,
  colframe=green!40!black,
  colbacktitle=green!10!white,
  coltitle=black,
  fonttitle=\bfseries\small,
  fontupper=\small,
  title={Definition~\thetcbcounter: #2},
  label={def:#2},
  attach boxed title to top left={yshift=-3.25mm, xshift=2mm},
  boxed title style={colframe=green!40!black},
  left=10pt,
  right=10pt,
  top=9pt,
  bottom=3pt,
  arc=1.5pt,
  drop shadow=black!25,
  #1
}

\newtcolorbox[auto counter, number within=section]{mythm}[2][]{%
  enhanced,
  breakable,
  colback=cyan!1, 
  colframe=cyan!40!black, 
  colbacktitle=cyan!10!white, 
  coltitle=black, 
  fonttitle=\bfseries\small,
  fontupper=\small, 
  title={Theorem~\thetcbcounter: #2}, % 手动计数
  label={thm:#2},
  attach boxed title to top left={yshift=-3.25mm, xshift=2mm},
  boxed title style={colframe=cyan!35!black}, 
  left=10pt, 
  right=10pt, 
  top=9pt, 
  bottom=3pt, 
  arc=1.5pt, 
  drop shadow=black!25, 
  #1
}

\BeforeBeginEnvironment{wrapfigure}{\setlength{\intextsep}{0pt}}
\BeforeBeginEnvironment{wraptable}{\setlength{\intextsep}{0pt}}

\usepackage{comment} 
\usepackage{enumitem}
\theoremstyle{remark}

\usepackage{placeins}
\usepackage{booktabs}
\usepackage{tabularx}

\usepackage{soul}
\definecolor{highlightyellow}{RGB}{255,193,7}   % amber border; tweak to taste

\newtcolorbox{diagnosticbox}[1][]{
  enhanced, breakable,
  colback=gradblue!2,
  colframe=gradblue!45,
  boxrule=0.45pt,
  leftrule=2.4pt,
  arc=2pt, outer arc=2pt,
  left=6pt, right=6pt, top=3pt, bottom=3pt,
  before skip=3pt, after skip=3pt,
  colback=white,
  colframe=black!16,
  boxrule=0.4pt,
  borderline west={2.6pt}{0pt}{gradblue},
  arc=2pt, outer arc=2pt,
  left=8pt, right=8pt, top=5pt, bottom=6pt,
  before skip=6pt, after skip=6pt,
  fontupper=\small,
  fonttitle=\bfseries\sffamily\small,
  coltitle=gradblue,
  colbacktitle=gradblue!9,
  coltitle=black,
  colbacktitle=white,
  title={#1},
  toptitle=2pt, bottomtitle=2pt,
  lefttitle=5pt, righttitle=5pt,
  segmentation style={solid, gradblue!20, line width=0.35pt},
  toptitle=5pt, bottomtitle=4pt,
  lefttitle=8pt, righttitle=8pt,
  titlerule=0.4pt,
  segmentation style={solid, black!12, line width=0.4pt},
}
\definecolor{gradblue}{RGB}{31,119,180}
\definecolor{noiseorange}{RGB}{214,96,27}
\definecolor{coveragepurple}{RGB}{117,107,177}
\definecolor{controlgreen}{RGB}{27,158,119}

\definecolor{citeblue}{HTML}{355C9A}
\definecolor{refteal}{HTML}{287D82}
\definecolor{urlbluegray}{HTML}{52677D}

\hypersetup{
    colorlinks=true,
    citecolor=citeblue,
    linkcolor=refteal,
    urlcolor=urlbluegray
}

\usepackage{mdframed}

\surroundwithmdframed[
  linewidth=0.5pt,
  linecolor=black,
  backgroundcolor=white,
  roundcorner=0pt,
  innerleftmargin=8pt,
  innerrightmargin=8pt,
  innertopmargin=6pt,
  innerbottommargin=6pt,
  skipabove=8pt,
  skipbelow=8pt
]{proposition}

\title{Understanding Negative Learning in Large Language Model Fine-tuning}

\author{Yue Wang$^{1,2}$\quad Qizhou Wang$^{3}$ \quad Zizhuo Zhang$^{2}$  \quad Gang Niu$^{3}$\quad Bo Han$^{1}$ \quad Masashi Sugiyama$^{3,4}$ \\ \\ $^1$ The University of Melbourne \quad $^2$ TMLR Group, Hong Kong Baptist University \\ $^3$ RIKEN Center for Advanced Intelligence Project  \quad $^4$ The University of Tokyo }

\iclrfinalcopy
\begin{document}

\etocdepthtag.toc{mtchapter}
\maketitle

\begin{abstract}
Negative learning has never been more important than it is today, particularly in large language model (LLM) fine-tuning, where it underpins machine unlearning, preference- and reward-based optimization, and exploratory data generation. Yet its functional role remains a subject of ongoing debate, with existing work variously characterizing it as forgetting, regularization, or target shaping. However, analytical support focuses primarily on the single-update scale, while empirical results at the trajectory scale are typically obtained under negative-only training and may not extend to composite objectives in which it interacts with other components. To develop a deeper understanding, we introduce a gradient-based endpoint probing framework that can examine negative updates separately throughout training. Our analysis reveals that negative learning plays different roles across algorithms and training phases. For example, in off-policy DPO, it acts more like a regularizer early in training and shows evidence of target shaping only toward out-of-distribution endpoints and only in later phases. In on-policy PPO and GRPO, by contrast, negative updates offset endpoint-aligned positive learning without stable target shaping, consistent with a role in sustaining exploration. Guided by these findings, we develop a family of stage-aware methods that modulate the strength of negative updates throughout training, improving fine-tuning performance across tasks.
\end{abstract}

\section{Introduction}
\emph{Negative learning}~\citep{welleck2019neural}, broadly defined as reducing the likelihood of selected outputs, is now widely used in large language model (LLM) fine-tuning. In machine unlearning~\citep{yao2024large,liu2025rethinking,wang2025towards}, it suppresses sensitive, harmful, or copyrighted content; in preference and policy optimization~\citep{ziegler2019fine,rafailov2023direct,zhang2024negative,shao2024deepseekmath}, it penalizes rejected responses or negative-advantage actions; and in controllable generation~\citep{lu2022quark,chen2026eepo}, it discourages undesirable or previously sampled responses to promote further exploration. Before its adoption in LLMs, related forms of negative supervision were studied in complementary-label, noisy-label, and partial-label learning~\citep{ishida2017learning,han2020sigua,lv2020progressive} to improve robustness and resolve target ambiguity.

Despite its widespread use, the {mechanisms} underlying negative learning remain considerably less understood. One prominent interpretation in LLM fine-tuning is \emph{implicit target shaping}. \citet{zhang2024negative} and \citet{zhu2026surprising} empirically show that, by suppressing given targets, negative-only training can shift learning toward more preferred alternatives. \citet{ren2025learning} analytically explain this effect by showing how a single negative-gradient update redistributes likelihood according to the model prior. These accounts accurately characterize negative learning when used in isolation. 
However, in its more common use within preference- and reward-optimization settings~\citep{ziegler2019fine,rafailov2023direct}, negative learning forms part of a composite objective and interacts with other components.
This interaction may alter the target-shaping behavior observed in isolation and cause the role of negative learning to evolve throughout training, leaving it unclear whether target shaping persists beyond a single update and remains stable to serve as a supervisory signal.
The picture is further complicated by alternative interpretations of negative learning as targeted forgetting~\citep{liu2025rethinking} or, more classically, as regularization~\citep{han2020sigua}. Consequently, its precise role within LLM fine-tuning remains difficult to isolate and understand.

To better understand negative learning, we need a flexible empirical tool that can disentangle its contributions from those of other components and characterize its role at the training-trajectory scale.
This is challenging because many fine-tuning objectives lack explicit learning targets against which individual components can be assessed. To address this issue, we introduce an \emph{endpoint probing} tool that treats responses generated by the final checkpoint as empirically achieved targets. Across training checkpoints, we quantify from a gradient perspective how strongly the current objective update supports these endpoints. Gradient additivity then allows each objective component to be probed separately at the same model state.
This construction builds upon established uses of gradient inner products~\citep{pruthi2020estimating,bengio2020interference}, including in prior work on LLMs~\citep{xia2024less,wang2025rethinking,chang2025scalable}. 
We further examine the agreement between endpoint probing estimates and actual one-step changes in endpoint likelihood in Section~\ref{sec:mechanism} and derive a local approximation-error bound in Appendix~\ref{app:t}, supporting the use of endpoint probing in our setting.

We deliberately keep our probing simple and do not overly emphasize its methodological novelty. 
Rather, we tailor this established technique to our setting, where it can reliably isolate and track component-wise effects that prior likelihood-based analyses~\citep{ren2025learning} cannot distinguish. Our  contribution lies in the empirical characterization of negative learning enabled by this framework.
We first confirm that negative-only learning can shape stable implicit targets, and then examine its role within composite objectives for preference and reward optimization.

We find that the role of negative learning within composite objectives varies with the optimization paradigm, endpoint distribution, and training phase. In off-policy preference optimization, exemplified by \emph{direct preference optimization} (DPO)~\citep{rafailov2023direct}, negative learning shapes targets for semantically related out-of-distribution prompts, but not for in-distribution prompts. This contrast does not imply a DPO deficiency. Rather, it suggests that negative target shaping may support preference generalization beyond the training distribution~\citep{im2024generalization}. 
Moreover, this behavior is phase-dependent, with negative learning acting more like a regularizer early in training and shaping the endpoints only in later phases.
By contrast, in on-policy reward optimization, represented by \emph{proximal policy optimization} (PPO)~\citep{ziegler2019fine} and \emph{group relative policy optimization} (GRPO)~\citep{shao2024deepseekmath}, we observe no stable target-shaping behavior from negative learning.
Instead, it offsets positive learning, consistent with a role in sustaining exploration, a central mechanism of reinforcement learning, toward alternative, potentially higher-reward responses.

These findings complement existing accounts of negative learning in LLM fine-tuning and may motivate further improvements to methods that rely on it. As a concrete demonstration, we develop stage-aware fine-tuning methods that adjust the relative strengths of positive and negative learning according to their evolving roles. These variants outperformed their corresponding baselines in preference alignment, reasoning, and tool use, showing how our findings can guide method design.

\section{Necessary Background}

We first introduce the background that motivates negative learning and its use in composite LLM fine-tuning objectives, and then revisit existing explanations of its functional role and their limitations. We keep the discussion concise and defer further details to Appendix~\ref{app:po method}.

\subsection{Positive, Negative, and Composite Learning}

Consider an LLM with conditional distribution $\pi(y\mid x;\boldsymbol{\theta})$ and a training set $\mathcal{D}^{+}=\{z^+=(x,y^{+})\}$, with $x$ the prompt, $y^{+}$ the target, and $\boldsymbol{\theta}$ the parameters. Supervised fine-tuning typically enables the model to learn from $\mathcal{D}^{+}$ by minimizing $\mathcal{L}^{+}(\boldsymbol{\theta})=-\hat{\mathbb{E}}_{\mathcal{D}^{+}}\log\pi(y^{+}\mid x;\boldsymbol{\theta})$, where $\hat{\mathbb{E}}$ denotes the empirical expectation. We refer to this likelihood-increasing process as \emph{positive learning}. However, modern fine-tuning often involves more complex objectives that require the model to move away from undesirable data, typically through \emph{negative learning}~\citep{yao2024large,liu2025rethinking}. Specifically, given a negative training set $\mathcal{D}^{-}=\{z^-=(x,y^{-})\}$, a canonical objective is $\mathcal{L}^{-}(\boldsymbol{\theta})=\hat{\mathbb{E}}_{\mathcal{D}^{-}}\log\pi(y^{-}\mid x;\boldsymbol{\theta})$, whose minimization decreases the likelihood of the selected outputs.

More practical LLM fine-tuning methods often require learning from desirable outputs while moving away from undesirable ones, thereby combining positive and negative learning. Objectives used in preference and reward optimization, such as DPO and PPO~\citep{rafailov2023direct,ziegler2019fine}, are prominent examples of what we call \emph{composite objectives}, written as
\begin{equation}
    \mathcal{L}_{\mathrm{comp}}(\boldsymbol{\theta})
    =\mathcal{L}^{+}(\boldsymbol{\theta})
    +\mathcal{L}^{-}(\boldsymbol{\theta}).
    \label{eq:composite-learning}
\end{equation}
This paper broadly defines positive and negative learning as objective components whose minimization increases and decreases the likelihood of selected outputs, respectively. Their specific formulations may incorporate data reweighting and other regularization strategies~\citep{wang2025rethinking,fan2026simplicity}. 
Please refer to Appendix~\ref{app:po method} for detailed formulations of DPO, PPO, and others.

\subsection{Understanding Negative Learning}

Our current understanding of positive learning is comparatively mature. Classical machine learning characterizes it through likelihood-based estimation, empirical risk minimization, and theories of learnability and generalization~\citep{bishop2006pattern,mohri2012foundations}. Recent work in LLMs has further studied how positive learning evolves during fine-tuning, linking its dynamics to memorization and knowledge acquisition and examining how task diversity, model scale, and data quality shape reasoning, instruction following, and safety~\citep{wei2021finetuned,sanh2021multitask,zhou2023lima,gekhman2024does,choi2024safety,pang2025token,wang2025breaking}.

By contrast, negative learning remains less understood because its canonical objective is unbounded below and specifies no learning target, complicating theoretical analyses of learnability and generalization. Consequently, empirical studies of LLM fine-tuning have focused on observed training behavior, showing that negative-only training can induce forgetting of negative examples~\citep{zhang2024negative} and shift the model toward implicit targets~\citep{zhu2026surprising}. \citet{ren2025learning} analytically characterize the latter behavior as probability squeezing, whereby a single negative update redistributes probability mass from the suppressed target toward alternatives favored under the current model distribution. This target-shaping account has since become influential, guiding subsequent analyses and method design in LLM fine-tuning~\citep{wang2026vpo,ren2026effect}.

We recognize that target shaping characterizes negative learning when studied in isolation, as we further verified in Section~\ref{sec:main_npo}, or analyzed locally at a single update.
However, extrapolating this local effect to explain the broader behavior of composite objectives in Eq.~\eqref{eq:composite-learning}, as \citet{ren2025learning} imply for DPO and PPO, can be misleading. As the model state evolves, interactions between negative learning and other components, particularly positive learning, may alter this behavior throughout training~\citep{wang2025gru,razin2025unintentional,yuan2025common}. Local target shaping therefore does not necessarily imply that negative learning shapes a stable target over the training trajectory. Such an incomplete account can obscure its actual contribution during joint training and misguide method design by treating a transient local effect as a persistent supervisory signal.

\section{Probing Mechanism}\label{sec:mechanism}
Motivated by this gap, we move beyond local characterization to trajectory-level verification, asking whether and when locally induced target shaping persists as a stable supervisory signal. This question is challenging because the implicit targets shaped by negative learning are not directly observable. Moreover, existing diagnostics, such as those that track likelihood changes~\citep{ren2025learning,razin2025unintentional,yan20253d}, capture the joint effects of positive and negative learning within composite objectives and therefore cannot isolate their respective contributions.

\subsection{Endpoint Construction}\label{sec:endpoint-construction}

To this end, we introduce \emph{endpoint probing} to examine whether locally induced target shaping persists throughout LLM fine-tuning. If a target consistently guides training, we expect its influence to be reflected in the final model responses. 
We therefore use responses generated by the final checkpoint as empirical surrogates for these otherwise unobservable targets and refer to them as \emph{endpoints}. Endpoint probing looks back along the training trajectory and, at each intermediate checkpoint, measures whether the current update moves the model toward or away from these endpoints. Tracking this effect across checkpoints provides evidence of whether target shaping toward these endpoints persists as a stable supervisory signal and when it emerges or disappears during training.
% These endpoints provide a concrete reference for assessing sustained target shaping, but a lack of alignment does not rule out shaping other implicit targets. In such cases, further analysis is needed to clarify its role.

Concretely, we construct an endpoint dataset $\mathcal{D}_{\mathrm{end}}=\{(x',y_{\mathrm{end}})\}$, where $x'$ is a held-out prompt and $y_{\mathrm{end}}$ is the corresponding response generated by the final model, parameterized by $\boldsymbol{\theta}^{(T)}$ after $T$ fine-tuning steps. The held-out prompts may be drawn from either the training distribution or a semantically related out-of-distribution task, allowing us to examine target shaping both within and beyond the training distribution.
We measure the model fit to the endpoints using the endpoint loss $\mathcal{L}_{\mathrm{end}}(\boldsymbol{\theta})=-\hat{\mathbb{E}}_{\mathcal{D}_{\mathrm{end}}}\log\pi(y_{\mathrm{end}}\mid x';\boldsymbol{\theta})$. 
Then, to probe the effect of an objective $\mathcal{L}$ at an intermediate checkpoint $\boldsymbol{\theta}^{(t)}$, we evaluate how this loss would change after one gradient update on $\mathcal{L}$:
\begin{equation}
\Delta(\mathcal{L};\boldsymbol{\theta}^{(t)}):=\mathcal{L}_{\mathrm{end}}(\boldsymbol{\theta}^{(t)})-\mathcal{L}_{\mathrm{end}}\big(\boldsymbol{\theta}^{(t)}-\eta\nabla_{\boldsymbol{\theta}}\mathcal{L}(\boldsymbol{\theta}^{(t)})\big),
\label{eq:endpoint-change}
\end{equation}
where $\eta>0$ is the learning rate. A positive $\Delta$ indicates that the update moves the model toward the endpoints, whereas a negative value indicates the opposite.

\subsection{Gradient-based Estimation}\label{sec:gradient-estimation}

Since $\mathcal{L}_{\mathrm{end}}$ is nonlinear, the exact change in Eq.~\eqref{eq:endpoint-change} for $\mathcal{L}=\mathcal{L}_{\mathrm{comp}}$ cannot generally be decomposed to isolate the contributions of positive and negative learning. Fortunately, gradients are additive, allowing component contributions to be separated under a first-order approximation.
This motivates the first-order expansion around $\boldsymbol{\theta}^{(t)}$, yielding $\Delta\approx\eta\nabla_{\boldsymbol{\theta}}\mathcal{L}_{\mathrm{end}}(\boldsymbol{\theta}^{(t)})^{\top}\nabla_{\boldsymbol{\theta}}\mathcal{L}(\boldsymbol{\theta}^{(t)})$. Omitting the positive scaling factor $\eta$, we define the \emph{endpoint-probing score} as
\begin{equation}
{\delta}(\mathcal{L};\boldsymbol{\theta}^{(t)}):=\nabla_{\boldsymbol{\theta}}\mathcal{L}_{\mathrm{end}}(\boldsymbol{\theta}^{(t)})^{\top}\nabla_{\boldsymbol{\theta}}\mathcal{L}(\boldsymbol{\theta}^{(t)}).
\label{eq:endpoint-probing}
\end{equation}
As in Eq.~\eqref{eq:endpoint-change}, $\delta>0$ predicts an update toward the endpoints, and $\delta<0$ the opposite.

\begin{wrapfigure}{r}{0.55\textwidth}
   \centering
    \begin{subfigure}[t]{0.27\textwidth}
        \centering
        \includegraphics[width=\textwidth]{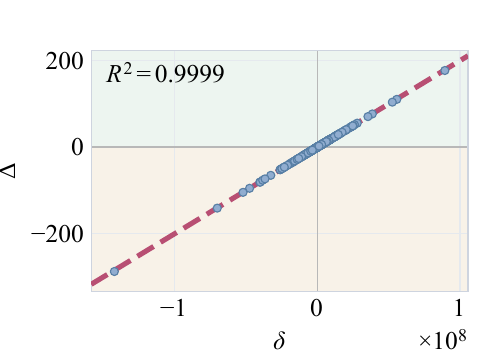}
        \caption{DPO on Qwen3-1.7B}
    \end{subfigure}
    \begin{subfigure}[t]{0.27\textwidth}
        \centering
        \includegraphics[width=\textwidth]{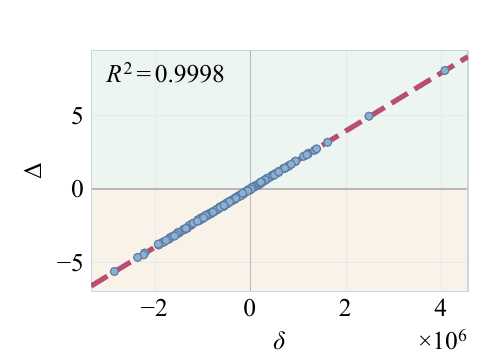}
        \caption{PPO on Pythia-2.8B}
    \end{subfigure}
    \caption{\textbf{Likelihood vs. Gradient Probing.} The endpoint-probing score $\delta$ strongly correlates with the exact one-step reduction in endpoint loss $\Delta$ across checkpoints, with squared Pearson correlation $R^2\approx1$. This supports using $\delta$ as a proxy for the update effect. {Details are provided in Appendix}~\ref{app:setup-probing}. }
    \label{fig:correlation}
\end{wrapfigure}
We derive the first-order approximation and its local error bound in Appendix~\ref{app:endpoint-approximation}. Under local smoothness assumptions, a small learning rate and gradient clipping help control the approximation error, both being standard practices in LLM fine-tuning~\citep{biderman2023pythia}. 
Thus, $\eta\delta$ provides a first-order approximation to $\Delta$,
with $\delta$ serving as a learning-rate-independent proxy.
To further validate this approximation empirically, we compared $\delta$ with the exact one-step change $\Delta$ across training checkpoints. Figure~\ref{fig:correlation} shows a strong positive correlation for both DPO and PPO, indicating that endpoint probing closely tracks the actual update effect. 

\section{Probing Results}\label{sec:results}

We applied endpoint probing to NPO, DPO, PPO, and GRPO~\citep{ziegler2019fine,zhang2024negative,shao2024deepseekmath} on unlearning, preference alignment, and tool-use reasoning, with Pythia, Qwen, Llama, and Gemma families~\citep{biderman2023pythia,grattafiori2024llama,yang2025qwen3,gemma4_2026}.  
Results were averaged over three independent trials, with more details in Appendix~\ref{app:exp_set}.

\subsection{Negative-only Optimization}
\label{sec:main_npo}
We begin with \emph{negative preference optimization} (NPO)~\citep{zhang2024negative}, a representative negative-only method for LLM unlearning. NPO learns only from undesirable prompt–response pairs, telling the LLM what to avoid without specifying a preferred alternative. It was developed to remove private or harmful information while preserving general model utility, and has since become a widely used baseline in LLM unlearning and inspired subsequent work~\citep{wang2025gru,wang2025rethinking,fan2026simplicity}.

The original NPO objective is relatively complex, so we use its more interpretable gradient-equivalent form. Given a forget set $\mathcal D_{\mathrm{npo}}=\{z^-=(x,y^-)\}$, NPO is gradient-equivalent to minimizing
\begin{equation}
\mathcal L^-_{\mathrm{npo}}(\boldsymbol\theta)=
\hat{\mathbb E}_{\mathcal D_{\mathrm{npo}}}
\big[
\omega_{\mathrm{npo}}
(z^-;\boldsymbol\theta\vert_{\mathrm{detach}})
\log\pi(y^-\mid x;\boldsymbol\theta)
\big].
\label{eq:npo}
\end{equation}
The weight $\omega_{\mathrm{npo}}$ depends on the current model $\boldsymbol\theta$ and negative example $z^-$ but is treated as constant during differentiation, as indicated by $\vert_{\mathrm{detach}}$. It downweights undesirable responses whose likelihood has already been substantially reduced relative to the reference model, thereby limiting excessive negative updates. The original objective and detailed derivation are provided in Appendix~\ref{sec:npo}.

\begin{wrapfigure}{r}{0.55\textwidth}
    \centering
    \begin{subfigure}[t]{0.27\textwidth}
        \centering
        \includegraphics[width=\textwidth]{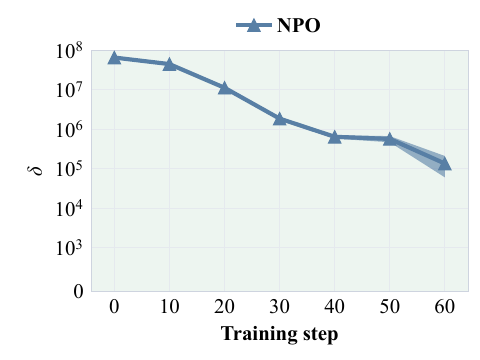}
        \caption{{Llama3.2-1B}}\label{fig:npo 1b}
    \end{subfigure}
    \begin{subfigure}[t]{0.27\textwidth}
        \centering
        \includegraphics[width=\textwidth]{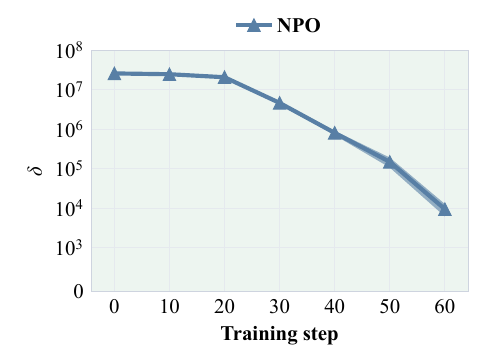}
        \caption{{Llama3.2-3B}}\label{fig:npo 3b}
    \end{subfigure}
    \caption{
    \textbf{NPO Probing.} We report mean endpoint-probing scores for NPO on TOFU \texttt{forget05} using Llama3.2-1B and 3B, with shading indicating the standard deviation.
    }
    \label{fig:npo-probing}
\end{wrapfigure}
We probed NPO on the TOFU \texttt{forget05} task~\citep{maini2024tofu} using Llama3.2-1B and 3B~\citep{grattafiori2024llama}. Training lasted 10 epochs (70 steps), with checkpoints saved every 10 steps.
This interval balances temporal resolution and computational efficiency, while Appendix~\ref{app:theory 1} gives conditions under which checkpoint scores remain informative between observations.
Figure~\ref{fig:npo-probing} presents the results. 
Although NPO specifies only what the LLM should avoid, its positive probing scores across checkpoints provide evidence of stable target shaping, consistent with findings on negative-only learning in LLM reasoning~\citep{zhu2026surprising}.

\subsection{Off-policy Preference Optimization}
Among off-policy methods with composite objectives, DPO~\citep{rafailov2023direct} is a widely used representative. It is typically used for preference optimization, learning from fixed preference pairs by increasing the relative likelihood of preferred responses over rejected ones, without online sampling or explicit reward-model training. Although the original DPO objective does not directly take the form of Eq.~\eqref{eq:composite-learning}, Appendix~\ref{sec:dpo} derives the following gradient-equivalent decomposition:
\begin{equation}
    \mathcal L_{\mathrm{dpo}}(\boldsymbol\theta)=\underbrace{\hat{\mathbb E}_{\mathcal D_{\mathrm{dpo}}}\big[-\omega_{\mathrm{dpo}}(z;\boldsymbol{\theta}{\vert_\mathrm{detach}})\log\pi(y^+|x;\boldsymbol\theta)\big]}_{\mathcal L_{\mathrm{dpo}}^+(\boldsymbol\theta)}
    +\underbrace{\hat{\mathbb E}_{\mathcal D_{\mathrm{dpo}}}\big[\omega_{\mathrm{dpo}}(z;\boldsymbol{\theta}{\vert_\mathrm{detach}})\log\pi(y^-|x;\boldsymbol\theta)\big]}_{\mathcal L_{\mathrm{dpo}}^-(\boldsymbol\theta)}.\label{eq:dpo}
\end{equation}
Here, $\mathcal D_{\mathrm{dpo}}=\{z=(x,y^+,y^-)\}$ is a dataset of preference pairs, where $y^+$ is preferred over $y^-$. The weight $\omega_{\mathrm{dpo}}$ emphasizes preference pairs with insufficient margins and downweights those already well learned.
Accordingly, $\mathcal L_{\mathrm{dpo}}^+$ and $\mathcal L_{\mathrm{dpo}}^-$ represent positive and negative learning, respectively.

\begin{figure*}[t]
    \centering
    \begin{subfigure}[t]{0.3\textwidth}
        \centering
        \includegraphics[width=\textwidth]{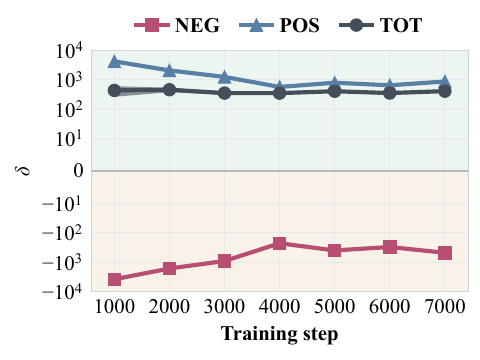}
        \caption{ID, Pythia}\label{fig:dpo id}
    \end{subfigure}\quad
    \begin{subfigure}[t]{0.3\textwidth}
        \centering
        \includegraphics[width=\textwidth]{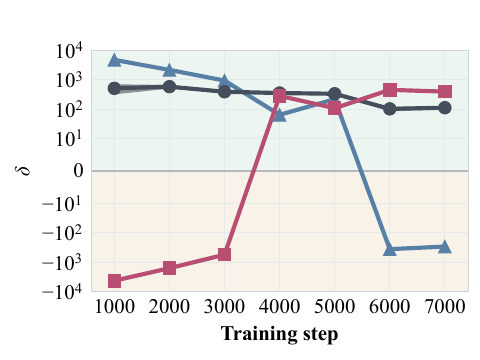}
        \caption{OOD, Pythia}\label{fig:dpo ood}
    \end{subfigure}\quad
    \begin{subfigure}[t]{0.3\textwidth}
        \centering
        \includegraphics[width=\textwidth]{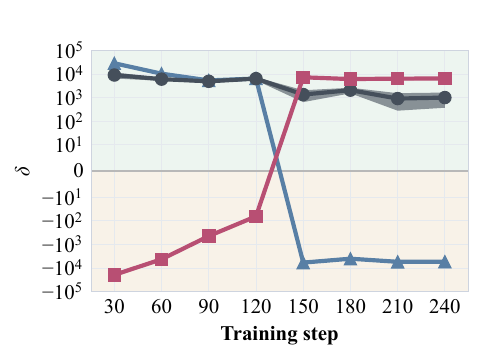}
        \caption{OOD, Qwen}\label{fig:dpo ood qwen}
    \end{subfigure}
    \caption{
    \textbf{DPO Probing.} We report endpoint-probing scores for in-distribution (ID) and out-of-distribution (OOD) endpoints using Pythia-2.8B (Pythia) and Qwen3-1.7B (Qwen). % We report scores for the full DPO objective  $\delta(\mathcal{L}_{\mathrm{dpo}};\boldsymbol{\theta})$, its positive component $\delta(\mathcal{L}_{\mathrm{dpo}}^{+};\boldsymbol{\theta})$, and its negative component $\delta(\mathcal{L}_{\mathrm{dpo}}^{-};\boldsymbol{\theta})$.
    }
    \label{fig:ablationa}
\end{figure*}

We probed DPO trained on UltraFeedback~\citep{cui2024ultrafeedback} using Pythia-2.8B~\citep{biderman2023pythia} as the base model. Following common practice~\citep{tunstall2023zephyr}, we performed supervised fine-tuning before DPO and tuned the hyper-parameters for stable training. We constructed the endpoint dataset by pairing held-out, in-distribution prompts with responses greedily decoded from the final DPO checkpoint. Figure~\ref{fig:dpo id} presents the results. $\mathcal L_{\mathrm{dpo}}$ yields positive scores across the probed checkpoints, indicating that DPO updates remain directed toward the endpoints and suggesting a relatively stable target. This pattern resembles supervised-like optimization dynamics, consistent with prior theoretical connections between supervised learning and DPO~\citep{wang2026implicit}. However, component-wise probing shows no clear target-shaping effect from negative learning on these endpoints, while positive learning primarily drives the model toward them.

Prior studies have reported that DPO improves generalization to related but unseen tasks~\citep{ranaldi2024self,wang2025direct}, motivating us to examine its target-shaping dynamics for out-of-distribution endpoints. We retained UltraFeedback for DPO but constructed endpoints from HH-RLHF-helpfulness prompts~\citep{bai2022training}, which cover similar helpfulness tasks but form a distinct prompt set. The probing results in Figure~\ref{fig:dpo ood} reveal intriguing dynamics. First, the full DPO objective remains positively aligned with the out-of-distribution endpoints across checkpoints, suggesting stable supervised-like dynamics beyond the training distribution. More importantly, the roles of its components evolve. Positive learning initially drives the model toward the endpoints while negative learning moves against them. Their roles later reverse, with negative learning shaping the endpoints and positive learning moving in the opposite direction. 

Figure~\ref{fig:dpo ood qwen} further shows a similar transition with Qwen3-1.7B~\citep{yang2025qwen3}, where negative learning exhibits sustained target shaping only in the later phase. This suggests that this behavior may extend across model families.
Negative learning shows no clear target-shaping effect on in-distribution endpoints but shapes out-of-distribution endpoints later in training. This contrast does not by itself indicate better or worse performance. Instead, it suggests that negative learning may contribute to the generalization advantage of DPO over supervised fine-tuning. 
More specifically, its target-shaping effect on unseen prompts offers a potential mechanism for this generalization benefit.

\begin{wrapfigure}{r}{0.55\textwidth}
   \centering
    \begin{subfigure}[t]{0.27\textwidth}
        \centering
        \includegraphics[width=\textwidth]{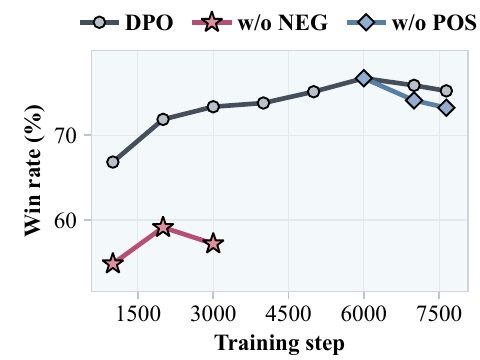}
        \caption{{{Pythia-2.8B}}}
    \end{subfigure}
    \begin{subfigure}[t]{0.27\textwidth}
        \centering
        \includegraphics[width=\textwidth]{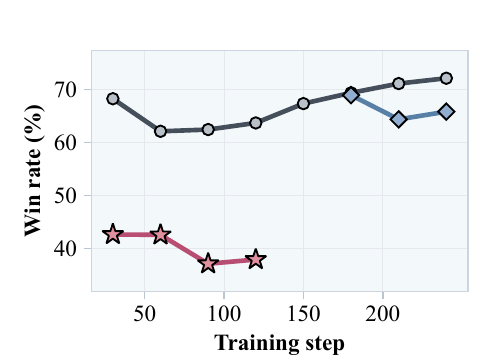}
        \caption{Qwen3-1.7B}
    \end{subfigure}
    \caption{\textbf{DPO Win-Rate Ablation.} 
    We report win rates after separately removing negative learning early and positive learning later, when each has a negative endpoint-probing score. These ablations suggest that components with negative $\delta$ can remain beneficial.
    }\label{fig:dpo ablation}
\end{wrapfigure}
What does a negative endpoint-probing score imply? By definition, it predicts that the component moves the model away from the endpoints. However, its effect on task performance, typically evaluated by win rate, remains ambiguous. 
It may harm performance by interfering with fine-tuning~\citep{yu2020gradient} or benefit it by regularizing training~\citep{han2020sigua}.
To distinguish these possibilities, we separately removed negative learning early and positive learning later, when their respective scores are negative. Figure~\ref{fig:dpo ablation} shows that both interventions reduce win rates, consistent with a beneficial regularization role rather than merely harmful interference. Specifically, negative learning constrains positive-driven target shaping early, while positive learning constrains negative-driven shaping later.

\begin{diagnosticbox}[Take-Home Message]
\begin{itemize}[leftmargin=1.2em, itemsep=2pt, topsep=0pt]
\item The full DPO objective remains aligned with the final endpoints, indicating stable, supervised-like optimization with implicit learning targets.
\item The target-shaping effect of negative learning is observed only for out-of-distribution endpoints, while in-distribution target shaping is primarily driven by positive learning.
\item For out-of-distribution endpoints, target shaping is phase-dependent, with positive learning dominating early and negative learning dominating later.
\item On out-of-distribution endpoints, positive and negative learning play complementary roles, alternating between target shaping and beneficial regularization-like effects.
\end{itemize}
\end{diagnosticbox}

\subsection{On-Policy Reward Optimization}
\label{sec:opro}
We next consider PPO~\citep{ziegler2019fine} and GRPO~\citep{shao2024deepseekmath}, two on-policy reinforcement learning methods. Both seek to maximize expected reward using responses sampled and evaluated during training. In practice, this reward-maximization problem is optimized through a clipped surrogate, which encourages higher likelihood for positive-advantage outputs and lower likelihood for negative-advantage outputs. The surrogate can therefore be decomposed into positive and negative learning. We begin with PPO, which we study in the context of preference alignment. For our component-wise analysis, its clipped policy surrogate admits the following decomposition:
\begin{equation}\begin{aligned}
\mathcal L_{\mathrm{ppo}}(\boldsymbol\theta)
&=\underbrace{\hat{\mathbb E}_{\mathcal D_{\mathrm{ppo}}}
\big[-\operatorname{clip}_{A(z)}\big(r(z;\boldsymbol\theta)\big)A(z)
\mathbbm 1_{\{A(z)>0\}}
\big]}_{\mathcal L_{\mathrm{ppo}}^{+}(\boldsymbol\theta)}
\\&\quad\quad\quad\quad\quad\quad\quad\quad+
\underbrace{\hat{\mathbb E}_{\mathcal D_{\mathrm{ppo}}}
\big[-\operatorname{clip}_{A(z)}\big(r(z;\boldsymbol\theta)\big)A(z)
\mathbbm 1_{\{A(z)<0\}}\big]}_{\mathcal L_{\mathrm{ppo}}^{-}(\boldsymbol\theta)}.
\label{eq:onpolicy-decomposition}
\end{aligned}\end{equation}
PPO assigns token-level advantages to the rollout dataset $\mathcal D_{\mathrm{ppo}}=\{z=((x,y_{<j}),y_j)\}$, where $(x,y)$ is a sampled trajectory, $j$ indexes the tokens in $y$, and $A(z)$ is the advantage assigned to token $y_j$.
The density ratio is $r(z;\boldsymbol\theta)=\frac{\pi(y_j\mid x,y_{<j};\boldsymbol\theta)}{\pi(y_j\mid x,y_{<j};\boldsymbol\theta_{\mathrm{old}})}$, where $\boldsymbol\theta_{\mathrm{old}}$ parameterizes the rollout policy from which $\mathcal D_{\mathrm{ppo}}$ is sampled. Although PPO uses a clipped importance-weighted surrogate rather than the direct log-likelihood weighting in Eqs.~\eqref{eq:npo} and~\eqref{eq:dpo}, $\mathcal L_{\mathrm{ppo}}^{+}$ and $\mathcal L_{\mathrm{ppo}}^{-}$ still implement positive and negative learning.
Please refer to Appendix~\ref{sec:ppo} for a more detailed discussion.

\begin{figure*}[t]
    \centering
    \begin{subfigure}[t]{0.3\textwidth}
        \centering
        \includegraphics[width=\textwidth]{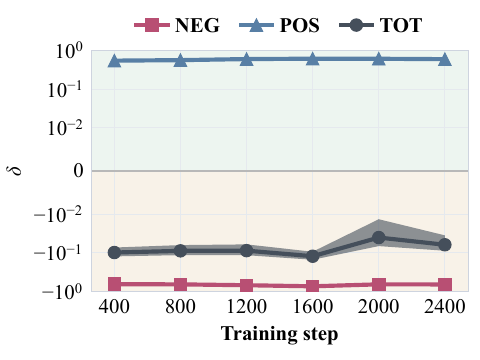}
        \caption{PPO, Pythia}\label{fig:ppo ood}
    \end{subfigure}\quad
    \begin{subfigure}[t]{0.3\textwidth}
        \centering
        \includegraphics[width=\textwidth]{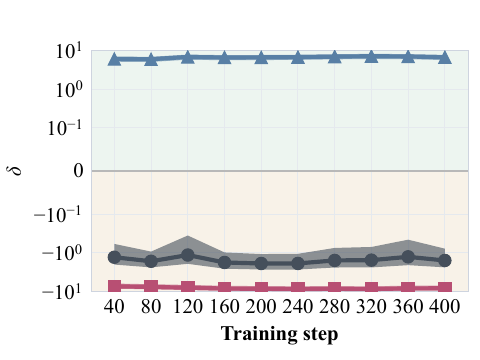}
        \caption{PPO, Qwen}\label{fig:ppo ood qwen}
    \end{subfigure}\quad
    \begin{subfigure}[t]{0.3\textwidth}
        \centering
        \includegraphics[width=\textwidth]{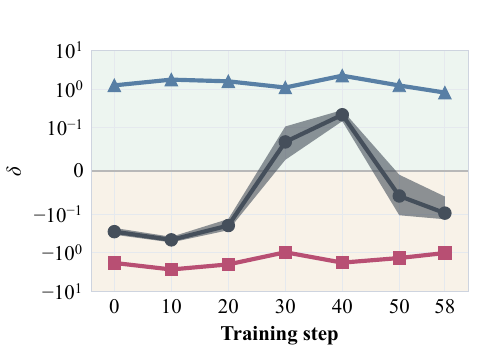}
        \caption{GRPO, Gemma}\label{fig:grpo}
    \end{subfigure}
    % \textbf{PPO and GRPO Probing.} We report endpoint-probing scores for OOD endpoints using Pythia-2.8B (Pythia) and Qwen3-1.7B (Qwen) for PPO, and Gemma4-E4B(Gemma) for GRPO. We report scores for the full PPO and GRPO objectives, $\delta(\mathcal{L}_{\mathrm{ppo}};\boldsymbol{\theta})$ and $\delta(\mathcal{L}_{\mathrm{grpo}};\boldsymbol{\theta})$, and their respective  components.
    \caption{
    \textbf{PPO and GRPO Probing.} We report probing scores using out-of-distribution endpoints for PPO with Pythia-2.8B (Pythia) and Qwen3-1.7B (Qwen), and held-out endpoints for GRPO with Gemma4-E4B (Gemma). 
    % We report for the full policy surrogates, $\delta(\mathcal L_{\mathrm{ppo}};\boldsymbol\theta)$ and $\delta(\mathcal L_{\mathrm{grpo}};\boldsymbol\theta)$, and their respective components.
    }
    \label{fig:ppo}
\end{figure*}

As in our DPO analysis, we used Pythia-2.8B and Qwen3-1.7B as base models and first performed supervised fine-tuning. We then trained the models with PPO on UltraFeedback, using the well tuned hyperparameters detailed in Appendix~\ref{app:exp_set}. Online rollouts were scored by ArmoRM-Llama3-8B-v0.1~\citep{ArmoRM}. We constructed the endpoint datasets by pairing held-out, out-of-distribution HH-RLHF-helpfulness prompts with responses greedily decoded from the final PPO checkpoints. 

\begin{wrapfigure}{r}{0.5\textwidth}
   \centering
    \begin{subfigure}[t]{0.245\textwidth}
        \centering
        \includegraphics[width=\textwidth]{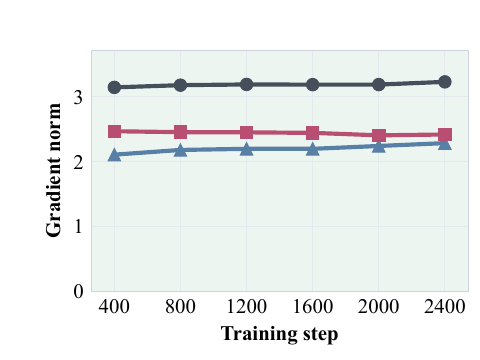}
        \caption{Gradient Norm}
    \end{subfigure}
    \begin{subfigure}[t]{0.245\textwidth}
        \centering
        \includegraphics[width=\textwidth]{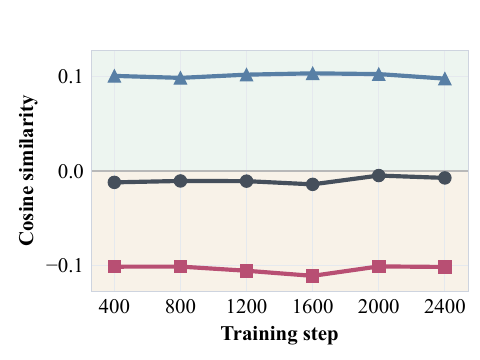}
        \caption{Cosine Similarity}
    \end{subfigure}
    \caption{
    \textbf{Gradient Norms and Cosine Similarity.} Using Pythia-2.8B under the same setup as Figure~\ref{fig:ppo ood}, we report gradient norms for the full PPO policy surrogate and its components, together with their cosine similarities with the endpoint gradient. % The full-surrogate gradient has a non-negligible norm but near-zero, slightly negative endpoint alignment.
    }\label{fig:ppo-normalized}
\end{wrapfigure}

The probing results can be found in Figures~\ref{fig:ppo ood}--\ref{fig:ppo ood qwen}.  Unlike DPO, PPO exhibits slightly negative, near-zero probing scores for both models, showing no stable endpoint-directed dynamics. Such scores may reflect either weak updates or nearly orthogonal gradients. 
To distinguish these cases, we report the gradient norms of the
PPO policy surrogate and its components, together with their cosine
similarities with the endpoint gradient, using Pythia-2.8B.
Figure~\ref{fig:ppo-normalized} shows that the full-surrogate gradient has a
non-negligible norm, while its cosine similarity with the
endpoint gradient remains close to zero and slightly negative. Thus, PPO gradients are mostly orthogonal to, but weakly opposed to, the endpoint gradients.
% This indicates active exploration with broad coverage of alternative directions, consistent with the on-policy reinforcement learning nature of PPO~\citep{ziegler2019fine}.
This pattern is consistent with exploration beyond the final responses, although endpoint alignment alone does not establish whether these updates discover better alternatives.

We next examine the roles of positive and negative learning. Across both models, the probing score of $\mathcal L_{\mathrm{ppo}}^{+}$ remains positive, indicating that updates from positive-advantage tokens move the model toward the final endpoints despite the continually evolving rollout dataset $\mathcal D_{\mathrm{ppo}}$. By contrast, the score of $\mathcal L_{\mathrm{ppo}}^{-}$ remains negative and slightly outweighs the positive component, resulting in a slightly negative score for the full PPO surrogate. Unlike in DPO, where negative learning eventually helps shape the endpoints, negative learning in PPO shows no stable target shaping at the trajectory scale and instead appears to sustain exploration over alternative responses. This finding complements the local account of target shaping through probability redistribution under individual negative updates, showing that the same local mechanism can play a different role across fine-tuning methods.

To examine whether this behavior also arises in other on-policy settings, we studied GRPO on reasoning and tool use. Unlike token-level PPO, GRPO assigns each sampled response a group-relative advantage obtained by normalizing rewards across responses to the same prompt. We trained Gemma4-E4B with GRPO on Hermes Function-Calling v1~\citep{quesnellehermes}, with details provided in Appendix~\ref{app:iidd_revw}. Figure~\ref{fig:grpo} shows a pattern similar to PPO. The full-surrogate score fluctuates around zero and is occasionally positive, but does not remain aligned with the final endpoints. The positive component is endpoint-aligned, whereas the negative component primarily offsets it. 
This suggests that the observed behavior extends beyond a single method or task.

\begin{diagnosticbox}[Take-Home Message]
\begin{itemize}[leftmargin=1.2em, itemsep=2pt, topsep=0pt]
\item Across the tasks studied, PPO and GRPO exhibit stable but opposing component-wise endpoint alignments, while the full-surrogate alignment may vary across settings.
\item Positive learning on positive-advantage tokens or responses consistently moves the model toward the endpoints despite evolving rollout data.
\item Negative learning counteracts this endpoint-directed tendency rather than consistently shaping the endpoints, consistent with a role in sustaining exploration over alternative responses.
\end{itemize}
\end{diagnosticbox}

\section{Controlled Negative Learning}
\label{sec:controlled-methods}

Our analyses reveal that the benefits of negative learning depend on both the involved method and the training phase. 
Negative learning shifts from a regularization-like role to target shaping for out-of-distribution endpoints during DPO, whereas, in PPO and GRPO, it consistently counteracts endpoint-aligned positive learning, suggesting a role in sustaining exploration.
These distinct dynamics suggest that negative learning should not be treated as uniformly beneficial or harmful. Instead, its influence could be coordinated with positive learning according to its method- and phase-specific role. Motivated by this finding, we introduce a family of \emph{controlled optimization} methods that adjust the relative strength of negative learning throughout training. The methods share a common principle but instantiate it differently for off-policy and on-policy optimization.

\subsection{Controlled DPO}
For the out-of-distribution endpoints studied, positive and negative learning exchange their target-shaping and regularization-like roles during DPO. This role exchange suggests placing greater relative emphasis on the component driving target shaping at each stage, while adjusting the regularization-like influence of the other component. We therefore introduce controlled DPO (cDPO): 
\begin{equation}
\mathcal L_{\mathrm{cdpo}}(\boldsymbol\theta)
=
2(1-\lambda_t)\mathcal L_{\mathrm{dpo}}^{+}(\boldsymbol\theta)
+
2\lambda_t\mathcal L_{\mathrm{dpo}}^{-}(\boldsymbol\theta).
\label{eq:controlled-dpo}
\end{equation}
$\lambda_t$ balances positive and negative learning, with $\lambda_t=0$, $1/2$, and $1$ recovering positive-only learning, standard DPO, and negative-only learning, respectively.
We set $\lambda_t=\max\big(\min(\frac{t-t_1}{t_2-t_1},1),0\big)$, where $t_1< t_2$ define a linear transition from positive to negative learning.

\subsection{Controlled PPO and GRPO}

Unlike in DPO, positive and negative learning exhibit stable but opposing endpoint alignments in the on-policy PPO and GRPO settings studied. Positive learning moves the model toward the final endpoints, suggesting that positive-advantage responses provide useful exploitation targets, while negative learning counteracts this direction. 
We examine whether controlling the strength of this counteracting effect can improve fine-tuning performance and therefore introduce controlled PPO (cPPO) and controlled GRPO (cGRPO), which share the following general form:
\begin{equation}
\mathcal L_{\mathrm{c}\mathsf{A}}(\boldsymbol\theta)
=
\mathcal L_{\mathsf{A}}^{+}(\boldsymbol\theta)
+\lambda_t\mathcal L_{\mathsf{A}}^{-}(\boldsymbol\theta),
\qquad \mathsf{A}\in\{\mathrm{ppo},\mathrm{grpo}\}.
\label{eq:controlled-ppo}
\end{equation}
$\lambda_t$ controls the strength of negative learning, with $\lambda_t=1$ recovering the standard policy surrogate and $\lambda_t<1$ and $\lambda_t>1$ attenuating and strengthening negative learning, respectively. Other terms remain unchanged.
We design three distinct scheduling rules for $\lambda_t$ to illustrate possible ways to coordinate positive and negative learning.
Periodic attenuation periodically weakens negative learning, allowing positive updates to more strongly reinforce currently high-reward responses.
Constant scaling maintains negative learning at a fixed reduced or enhanced strength, aiming to favor exploitation or exploration, respectively.
Adaptive strengthening, designed specifically for GRPO, increases the strength of negative learning as the fraction of fully successful rollout groups rises, aiming to encourage further exploration.
Please refer to Appendix~\ref{app:setup-controlled} for more details.

% We evaluate two scheduling rules for $\lambda_t$ (see Appendix~\ref{app:setup-controlled}): periodic attenuation, which periodically weakens negative learning to allow positive updates to learn more thoroughly from currently high-reward responses, and constant scaling, which maintains negative learning at a fixed reduced or enhanced strength to favor exploitation or exploration, respectively. 

\begin{figure*}[t]
    \centering
    \begin{subfigure}[t]{0.32\textwidth}
        \centering
        \includegraphics[width=\textwidth]{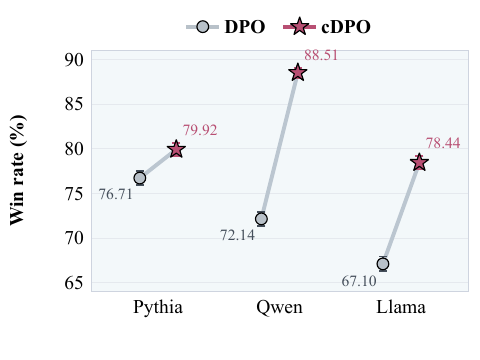}
        \caption{{DPO vs.\ cDPO}}\label{fig:po family dpo}
    \end{subfigure}\hfill
    \begin{subfigure}[t]{0.32\textwidth}
        \centering
        \includegraphics[width=\textwidth]{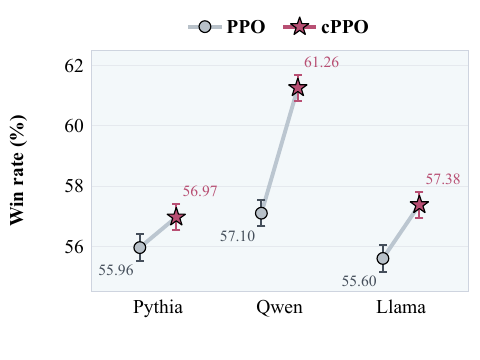}
        \caption{PPO vs.\ cPPO}\label{fig:po family ppo}
    \end{subfigure}\hfill
    \begin{subfigure}[t]{0.32\textwidth}
        \centering
        \includegraphics[width=\textwidth]{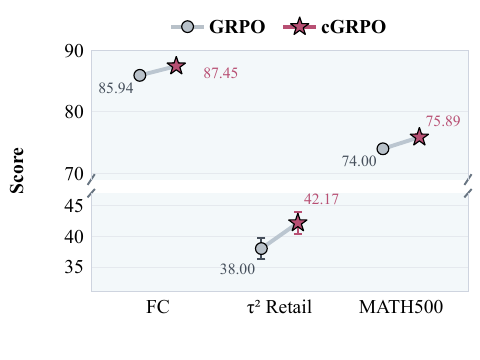}
        \caption{GRPO vs.\ cGRPO}\label{fig:po family grpo}
    \end{subfigure}
    \caption{\textbf{Controlled Variants.} 
    {Comparison of DPO, PPO, GRPO, and their controlled variants across models and tasks.} Markers indicate mean test scores across three independent training seeds, and error bars show $\pm$ one standard deviation across seeds.
    Metrics are AlpacaEval win rate against the corresponding supervised fine-tuning baseline for DPO and PPO, and mean programmatic reward (multiplied by 100), task-success rate (\%), and pass@1 (\%) for GRPO on function calling, $\tau^2$ Retail, and MATH500, respectively. % Hyperparameters and checkpoints are selected using held-out validation data, as described in Appendix~\ref{app:coordinate}.
    }
\label{fig:po family}
\end{figure*}

\subsection{Experimental Results} 

We evaluated cDPO, cPPO, and cGRPO against their standard counterparts across preference alignment, reasoning, and tool use. For preference alignment, we performed supervised fine-tuning on UltraChat-200k~\citep{ding2023enhancing}, trained DPO and PPO on UltraFeedback, and evaluated win rates on AlpacaEval~\citep{li2023alpacaeval}. In addition to the models used in our probing analysis, we further included Llama3-8B to examine whether the gains extend to a larger model. 
For reasoning and tool use, we compared GRPO and cGRPO on Hermes Function-Calling v1 using Gemma4-E4B, $\tau^2$ Retail~\citep{barres2025tau} using Qwen3.5-9B, and MATH500~\citep{lightman2024let} using Qwen3-0.6B.
Hyperparameters and checkpoints were selected on held-out validation data.

Figure~\ref{fig:po family} compares the controlled methods with their corresponding baselines using AlpacaEval win rate for DPO and PPO, and mean programmatic reward, task success rate, and pass@1 for GRPO on function calling, $\tau^2$ Retail, and MATH500, respectively. All experiments were repeated across three independent trials, and we report the mean and standard deviation.
The controlled methods achieve higher test scores across the evaluated models and tasks, suggesting that coordinating positive and negative learning can improve fine-tuning performance.
This may motivate further research on more dedicated and sophisticated coordination rules. Detailed experimental settings are provided in Appendix~\ref{app:exp_set}, and additional results can be found in Appendix~\ref{app:coordinate}.

\section{Conclusion}
Negative learning in LLM fine-tuning is often understood as implicit target shaping, but this account is incomplete. Our findings confirm stable target shaping under negative-only learning, yet reveal method- and phase-dependent roles within composite objectives. In DPO, negative learning shapes out-of-distribution endpoints only later in training and exhibits a regularization-like role early on. 
In PPO and GRPO, negative learning does not exhibit sustained target shaping toward the final-model endpoints. Instead, it consistently counteracts endpoint-aligned positive learning, consistent with a role in sustaining exploration.
Guided by this understanding, our controlled variants improved performance across the evaluated methods and tasks, highlighting the value of coordinating positive and negative learning according to their evolving roles and suggesting a direction for future research.

% \section*{AI Use Statement}
% We used LLM-based tools to polish portions of the manuscript, search relevant literature, and assist with mathematical derivations and proof checking. The authors take full responsibility for the accuracy, originality, and integrity of the paper, including its claims, citations, and conclusions.

% \section*{Ethics Statement}
% This work studies negative learning in LLM fine-tuning using established datasets and benchmarks. Potential risks include unintended changes to model behavior and degradation of capabilities outside the intended training scope. However, we do not identify particular ethical concerns specific to this study beyond those generally associated with LLM research and deployment.

% \section*{Reproducibility Statement}
% We used publicly available datasets and code libraries and provided our implementation in an anonymous repository, which will be released upon acceptance. We provided the hardware environments, experimental configurations, model choices, and hyperparameter settings in Appendix~\ref{app:exp_set}. Together, these resources and details reflect our best efforts to facilitate reproduction of our results.
\bibliography{ref}
\bibliographystyle{iclr2027_conference}
\clearpage

\appendix

% Standalone replacement draft for Appendix Experimental Configurations.
% Requires xcolor. Red text marks missing information or conflicts to resolve.
% Existing citation keys are retained where known; unverified references are
% explicitly marked in red rather than introduced as unresolved citations.
% This is an appendix fragment, not a standalone LaTeX document.

\section{LLM Fine-tuning Methods}
\label{app:po method}
In this section, we review NPO, DPO, PPO, GRPO, and on-policy distillation (OPD)~\citep{agarwal2024onpolicy}, and derive the objective forms used in our component-wise analysis.

\subsection{NPO}
\label{sec:npo}
Given a forget set $\mathcal D_{\mathrm{npo}}=\{z^-=(x,y^-)\}$, a simple approach to LLM unlearning minimizes the log-likelihood of the forget responses, which is given by $\mathcal L_{\mathrm{ga}}^-(\boldsymbol\theta)
=
\hat{\mathbb E}_{\mathcal D_{\mathrm{npo}}}
\left[
\log\pi(y^-\mid x;\boldsymbol\theta)
\right]$.
This is equivalent to gradient ascent on the negative log-likelihood. However, this objective is unbounded below and continues to encourage suppression even when the response likelihood is already very small, which can lead to unstable training and catastrophic model collapse~\citep{zhang2024negative,wang2025rethinking,wang2025towards}.

To mitigate this issue, NPO adopts a DPO-inspired objective using only negative examples, without requiring preferred alternatives.  Here, we consider the basic NPO objective without additional retention regularization, which is given by
\begin{equation}
\mathcal L_{\mathrm{npo}}^-(\boldsymbol\theta)
=
-\frac{2}{\beta}
\hat{\mathbb E}_{\mathcal D_{\mathrm{npo}}}
\Big[
\log\sigma\big(
-\beta\log
\frac{\pi(y^-\mid x;\boldsymbol\theta)}
{\pi(y^-\mid x;\boldsymbol\theta^{\mathrm{ref}})}
\big)
\Big],
\label{eq:npo-original}
\end{equation}
where $\boldsymbol\theta^{\mathrm{ref}}$ denotes the fixed reference-model parameters, $\beta>0$ controls the weighting behavior, and $\sigma$ is the sigmoid function. Unlike $\mathcal L_{\mathrm{ga}}$, this objective is bounded below by zero.
Using $\frac{\mathrm d}{\mathrm du}\log\sigma(u)=\sigma(-u)$, differentiation gives $\nabla_{\boldsymbol\theta}\mathcal L_{\mathrm{npo}}^-(\boldsymbol\theta)
=
\hat{\mathbb E}_{\mathcal D_{\mathrm{npo}}}
\left[
\omega_{\mathrm{npo}}(z^-;\boldsymbol\theta)
\nabla_{\boldsymbol\theta}
\log\pi(y^-\mid x;\boldsymbol\theta)
\right]$,
where
\begin{equation}
\omega_{\mathrm{npo}}(z^-;\boldsymbol\theta)
=
2\sigma\Big(
\beta\log
\frac{\pi(y^-\mid x;\boldsymbol\theta)}
{\pi(y^-\mid x;\boldsymbol\theta^{\mathrm{ref}})}
\Big).
\label{eq:npo-weight}
\end{equation}
The weight decreases toward zero as the response likelihood is reduced relative to the reference model, attenuating further negative updates on already suppressed responses.
Treating this weight as constant during differentiation yields the gradient-equivalent form used in Eq.~\eqref{eq:npo}:
\begin{equation}
\hat{\mathcal L}_{\mathrm{npo}}^{-}(\boldsymbol\theta)
=
\hat{\mathbb E}_{\mathcal D_{\mathrm{npo}}}
\left[
\omega_{\mathrm{npo}}
(z^-;\boldsymbol\theta\vert_{\mathrm{detach}})
\log\pi(y^-\mid x;\boldsymbol\theta)
\right].
\label{eq:npo-surrogate}
\end{equation}
Here, $\vert_{\mathrm{detach}}$ indicates that $\omega_{\mathrm{npo}}$ is treated as constant during differentiation. Thus,
$\nabla_{\boldsymbol\theta}\mathcal L_{\mathrm{npo}}^{-}
=\nabla_{\boldsymbol\theta}\hat{\mathcal L}_{\mathrm{npo}}^{-}$,
although their objective values need not coincide. For simplicity, Eq.~\eqref{eq:npo} in the main text denotes this gradient-equivalent surrogate by $\mathcal L_{\mathrm{npo}}^{-}$.

\subsection{DPO}
\label{sec:dpo}
DPO starts from KL-regularized reward maximization, whose optimal policy takes the form of $\pi^{\ast}(y\mid x)=\frac{1}{Z(x)}\pi(y\mid x;\boldsymbol\theta^{\mathrm{ref}})
\exp\Big(\frac{R(x,y)}{\beta}\Big)$, where $Z(x)$ is the partition function.  Under the Bradley--Terry preference model~\citep{bradley1952rank}, the associated preference probability is given by
$p(y^+\succ y^-\mid x)=\sigma\Big( \beta\log\frac{\pi^\ast(y^+\mid x)}{\pi(y^+\mid x;\boldsymbol\theta^{\mathrm{ref}})}-\beta\log\frac{\pi^\ast(y^-\mid x)}{\pi(y^-\mid x;\boldsymbol\theta^{\mathrm{ref}})}\Big)$.
DPO then minimizes the negative log-likelihood of the observed preferences, yielding its canonical objective:
\begin{equation}
\mathcal L_{\mathrm{dpo}}(\boldsymbol\theta)
=
-\hat{\mathbb E}_{\mathcal D_{\mathrm{dpo}}}
\Big[
\log\sigma\Big(
\beta\Big[
\log\frac{\pi(y^+\mid x;\boldsymbol\theta)}
{\pi(y^+\mid x;\boldsymbol\theta^{\mathrm{ref}})}
-
\log\frac{\pi(y^-\mid x;\boldsymbol\theta)}
{\pi(y^-\mid x;\boldsymbol\theta^{\mathrm{ref}})}
\Big]
\Big)
\Big].
\label{eq:dpo1}
\end{equation}

To make the component roles explicit, we present a gradient-equivalent form:
\begin{equation}
\hat{\mathcal L}_{\mathrm{dpo}}(\boldsymbol\theta)
=
-\hat{\mathbb E}_{\mathcal D_{\mathrm{dpo}}}
\left[
\omega_{\mathrm{dpo}}(z;\boldsymbol\theta\vert_{\mathrm{detach}})
\left(
\log\pi(y^+\mid x;\boldsymbol\theta)
-
\log\pi(y^-\mid x;\boldsymbol\theta)
\right)
\right],
\label{eq:dpo2}
\end{equation}
where the preference weight is
\begin{equation}
\omega_{\mathrm{dpo}}(z;\boldsymbol\theta)
=
\beta\sigma\Big(
-\beta\log\frac{\pi(y^+\mid x;\boldsymbol\theta)}
{\pi(y^-\mid x;\boldsymbol\theta)}
+
\beta\log\frac{\pi(y^+\mid x;\boldsymbol\theta^{\mathrm{ref}})}
{\pi(y^-\mid x;\boldsymbol\theta^{\mathrm{ref}})}
\Big).
\end{equation}
Treating the weight as constant during differentiation yields $\nabla_{\boldsymbol\theta}\mathcal L_{\mathrm{dpo}}=\nabla_{\boldsymbol\theta}\hat{\mathcal L}_{\mathrm{dpo}}$.
Also, Eq.~\eqref{eq:dpo} in the main text denotes this gradient-equivalent surrogate by $\mathcal L_{\mathrm{dpo}}$.
Accordingly, the positive- and negative-learning components are
\begin{align}
\mathcal L_{\mathrm{dpo}}^{+}(\boldsymbol\theta)
&=
-\hat{\mathbb E}_{\mathcal D_{\mathrm{dpo}}}
\left[
\omega_{\mathrm{dpo}}(z;\boldsymbol\theta\vert_{\mathrm{detach}})
\log\pi(y^+\mid x;\boldsymbol\theta)
\right],
\label{eq:dpo-p}\\
\mathcal L_{\mathrm{dpo}}^{-}(\boldsymbol\theta)
&=
\hat{\mathbb E}_{\mathcal D_{\mathrm{dpo}}}
\left[
\omega_{\mathrm{dpo}}(z;\boldsymbol\theta\vert_{\mathrm{detach}})
\log\pi(y^-\mid x;\boldsymbol\theta)
\right].
\label{eq:dpo-n}
\end{align}
Minimizing these components encourages increasing the likelihood of $y^+$ and decreasing that of $y^-$, respectively.
Moreover, we examined online DPO~\citep{guo2024direct} in Section~\ref{sec:online-dpo}.
Unlike the fixed preference dataset used in standard DPO, online DPO generates new response pairs from the current policy during training and assigns their preference order using a reward model. For each collected batch, the DPO objective and its gradient-equivalent decomposition above apply in the same way, with the sampled pairs held fixed during the update.

\subsection{PPO}
\label{sec:ppo}
PPO seeks to directly maximize the expected reward
$\mathcal J(\boldsymbol\theta)=\hat{\mathbb E}_{\mathcal D}\mathbb E_{\pi_{\boldsymbol\theta}}[R(x,y)]$, where the expectations average over training prompts and responses sampled from the current policy, respectively. Applying the likelihood-ratio identity~\citep{williams1992simple} gives
$\nabla_{\boldsymbol\theta}\mathcal J(\boldsymbol\theta)=\hat{\mathbb E}_{\mathcal D}\mathbb E_{\pi_{\boldsymbol\theta}}\left[R(x,y)\nabla_{\boldsymbol\theta}\log\pi(y\mid x;\boldsymbol\theta)\right]$.
Expanding the response log-likelihood across tokens and expressing expected returns relative to a state-value baseline gives the equivalent token-level expression
\begin{equation}
\nabla_{\boldsymbol\theta}\mathcal J(\boldsymbol\theta)
=
\hat{\mathbb E}_{\mathcal D}
\mathbb E_{\pi_{\boldsymbol\theta}}
\Big[
\sum_{j=1}^{|y|}
A^{\pi_{\boldsymbol\theta}}(x,y_{<j},y_j)
\nabla_{\boldsymbol\theta}
\log\pi(y_j\mid x,y_{<j};\boldsymbol\theta)
\Big],
\end{equation}
where $A^{\pi_{\boldsymbol\theta}}$ measures the expected return of a token relative to the state-value baseline.

PPO reuses responses sampled from an old policy parameterized by $\boldsymbol\theta_{\mathrm{old}}$. Following the main text, we denote the tokenized rollout dataset by $\mathcal D_{\mathrm{ppo}}=\{z=((x,y_{<j}),y_j)\}$ and the estimated advantage of token $y_j$ by $A(z)$. The advantages are held fixed during each policy-optimization phase. Using the token-level importance ratio
$r(z;\boldsymbol\theta)
=
\frac{\pi(y_j\mid x,y_{<j};\boldsymbol\theta)}
{\pi(y_j\mid x,y_{<j};\boldsymbol\theta_{\mathrm{old}})}$,
PPO constructs the local surrogate
$\mathcal J_{\mathrm{sur}}(\boldsymbol\theta)
=
\hat{\mathbb E}_{\mathcal D_{\mathrm{ppo}}}
\left[
r(z;\boldsymbol\theta)A(z)
\right]$.
Its gradient is $\nabla_{\boldsymbol\theta}\mathcal J_{\mathrm{sur}}(\boldsymbol\theta)
=
\hat{\mathbb E}_{\mathcal D_{\mathrm{ppo}}}
\left[
r(z;\boldsymbol\theta)A(z)
\nabla_{\boldsymbol\theta}
\log\pi(y_j\mid x,y_{<j};\boldsymbol\theta)
\right]$,
which provides a local policy-gradient approximation using the fixed rollout data.

To discourage excessively large policy changes, PPO clips the surrogate in the direction favored by the advantage. Written as a loss to minimize, its clipped policy surrogate is
\begin{equation}
\mathcal L_{\mathrm{ppo}}(\boldsymbol\theta)
=
-\hat{\mathbb E}_{\mathcal D_{\mathrm{ppo}}}
\left[
\operatorname{clip}_{A(z)}
\bigl(r(z;\boldsymbol\theta)\bigr)A(z)
\right],
\label{eq:ppo}
\end{equation}
where
\begin{equation}
\operatorname{clip}_{a}(r)
=
\begin{cases}
\min(r,1+\epsilon), & a\geq 0,\\
\max(r,1-\epsilon), & a<0,
\end{cases}
\end{equation}
and $\epsilon>0$ is the clipping threshold. 
The sign of the advantage therefore yields the exact decomposition
\begin{align}
\mathcal L_{\mathrm{ppo}}^{+}(\boldsymbol\theta)
&=
-\hat{\mathbb E}_{\mathcal D_{\mathrm{ppo}}}
\left[
\operatorname{clip}_{A(z)}
\bigl(r(z;\boldsymbol\theta)\bigr)A(z)
\mathbbm{1}_{\{A(z)>0\}}
\right],
\label{eq:ppo-p}\\
\mathcal L_{\mathrm{ppo}}^{-}(\boldsymbol\theta)
&=
-\hat{\mathbb E}_{\mathcal D_{\mathrm{ppo}}}
\left[
\operatorname{clip}_{A(z)}
\bigl(r(z;\boldsymbol\theta)\bigr)A(z)
\mathbbm{1}_{\{A(z)<0\}}
\right].
\label{eq:ppo-n}
\end{align}
Outside the clipped regions, minimizing the two components encourages increasing the likelihood of positive-advantage tokens and decreasing that of negative-advantage tokens, respectively. 
Throughout this decomposition, $\mathcal L_{\mathrm{ppo}}$ denotes the clipped policy-surrogate term rather than the complete training loss, which may additionally include explicit KL terms outside the scope of our analysis.

\subsection{GRPO}
\label{sec:grpo}
GRPO follows the policy-surrogate approach of PPO but estimates advantages from group-relative rewards without a learned value model. We consider its outcome-supervised form, where each sampled response receives a scalar reward.
For each prompt $x$, the old policy generates $G$ responses $\{y_i\}_{i=1}^{G}$. Their group-relative advantages are
\begin{equation}
A(x,y_i)
=
\frac{
R(x,y_i)-\frac{1}{G}\sum_{k=1}^{G}R(x,y_k)
}{
\operatorname{std}\bigl(\{R(x,y_k)\}_{k=1}^{G}\bigr)
},
\label{eq:grpo-advantage}
\end{equation}
where $\operatorname{std}$ denotes the standard deviation of the rewards within each group.
For groups with zero reward variance, we define all advantages as zero. Each response shares its advantage across all optimized tokens, and the advantages remain fixed during policy optimization.

We denote the response-level rollout dataset by
$\mathcal D_{\mathrm{grpo}}=\{z=(x,y)\}$,
where each response retains the advantage computed within its sampling group. The empirical expectation averages over prompts and then equally over the $G$ responses generated for each prompt. Although advantages are response-level, the importance ratio remains token-wise as in PPO, which is given by $r_j(z;\boldsymbol\theta)=\frac{\pi(y_j\mid x,y_{<j};\boldsymbol\theta)}{\pi(y_j\mid x,y_{<j};\boldsymbol\theta_{\mathrm{old}})}$.
Using the clipping function defined in Appendix~\ref{sec:ppo}, the clipped policy surrogate is
\begin{equation}
\mathcal L_{\mathrm{grpo}}(\boldsymbol\theta)
=
-\hat{\mathbb E}_{\mathcal D_{\mathrm{grpo}}}
\Big[
\frac{1}{|y|}
\sum_{j=1}^{|y|}
\operatorname{clip}_{A(z)}
\bigl(r_j(z;\boldsymbol\theta)\bigr)A(z)
\Big].
\label{eq:grpo}
\end{equation}
The within-response average preserves the response-level weighting rather than assigning greater weight to longer responses.
As in PPO, the sign of the advantage yields an exact decomposition:
\begin{align}
\mathcal L_{\mathrm{grpo}}^{+}(\boldsymbol\theta)
&=
-\hat{\mathbb E}_{\mathcal D_{\mathrm{grpo}}}
\Big[
\frac{1}{|y|}
\sum_{j=1}^{|y|}
\operatorname{clip}_{A(z)}
\bigl(r_j(z;\boldsymbol\theta)\bigr)A(z)
\mathbbm{1}_{\{A(z)>0\}}
\Big],
\label{eq:grpo-p}\\
\mathcal L_{\mathrm{grpo}}^{-}(\boldsymbol\theta)
&=
-\hat{\mathbb E}_{\mathcal D_{\mathrm{grpo}}}
\Big[
\frac{1}{|y|}
\sum_{j=1}^{|y|}
\operatorname{clip}_{A(z)}
\bigl(r_j(z;\boldsymbol\theta)\bigr)A(z)
\mathbbm{1}_{\{A(z)<0\}}
\Big].
\label{eq:grpo-n}
\end{align}
Positive and negative learning operate on responses with rewards above and below the group mean, respectively. 
As in PPO, $\mathcal L_{\mathrm{grpo}}$ denotes the clipped policy-surrogate term. The additional reference-policy KL penalty lies outside the scope of this component-wise analysis.
% Then, cGRPO follows directly: $\mathcal L_{\mathrm{cgrpo}}(\boldsymbol\theta)=\mathcal L_{\mathrm{grpo}}^{+}(\boldsymbol\theta)+\lambda_t\mathcal L_{\mathrm{grpo}}^{-}(\boldsymbol\theta)$, where $\lambda_t$ follows the same schedule as discussed in Section~\ref{sec:mechanism} and Appendix~\ref{app:complete-probing} to control the strength of negative learning.

\subsection{OPD}
\label{app:opd-objective}

OPD aims to transfer the capabilities of a stronger teacher to a student model by training on student-generated responses with supervision from the frozen teacher. Unlike PPO and GRPO, its learning signal comes from the teacher distribution rather than task rewards. Here, we consider the sampled-token formulation.
Following PPO, we denote the tokenized rollout dataset by
$\mathcal D_{\mathrm{opd}}=\{z=((x,y_{<j}),y_j)\}$.
The teacher--student log-probability gap defines the token-level
learning signal:
\begin{equation}
A(z;\boldsymbol\theta)
=
\log \pi_{\mathrm T}(y_j\mid x,y_{<j})
-
\log \pi(y_j\mid x,y_{<j};\boldsymbol\theta),
\label{eq:opd-gap}
\end{equation}
where $\pi_{\mathrm T}$ denotes the frozen teacher.
Positive and negative values indicate that the teacher assigns
higher and lower probability to the sampled token than the student,
respectively. The learning signal is held fixed during differentiation.
Using the token-level importance ratio $r(z;\boldsymbol\theta)$
and clipping function defined for PPO, the OPD policy surrogate is given by
\begin{equation}
\mathcal L_{\mathrm{opd}}(\boldsymbol\theta)
=
-\hat{\mathbb E}_{\mathcal D_{\mathrm{opd}}}
\left[
\operatorname{clip}_{A(z;\boldsymbol\theta)}
\bigl(r(z;\boldsymbol\theta)\bigr)
A(z;\boldsymbol\theta|_{\mathrm{detach}})
\right].
\label{eq:opd-surrogate}
\end{equation}
The empirical expectation averages over all valid completion tokens
in the rollout batch, preserving token-level rather than
response-level weighting.

As in PPO and GRPO, the sign of the learning signal yields
an exact decomposition:
\begin{align}
\mathcal L_{\mathrm{opd}}^{+}(\boldsymbol\theta)
&=
-\hat{\mathbb E}_{\mathcal D_{\mathrm{opd}}}
\left[
\operatorname{clip}_{A(z;\boldsymbol\theta)}
\bigl(r(z;\boldsymbol\theta)\bigr)
A(z;\boldsymbol\theta|_{\mathrm{detach}})
\mathbf{1}_{\{A(z;\boldsymbol\theta)>0\}}
\right],
\label{eq:opd-positive}
\\
\mathcal L_{\mathrm{opd}}^{-}(\boldsymbol\theta)
&=
-\hat{\mathbb E}_{\mathcal D_{\mathrm{opd}}}
\left[
\operatorname{clip}_{A(z;\boldsymbol\theta)}
\bigl(r(z;\boldsymbol\theta)\bigr)
A(z;\boldsymbol\theta|_{\mathrm{detach}})
\mathbf{1}_{\{A(z;\boldsymbol\theta)<0\}}
\right].
\label{eq:opd-negative}
\end{align}
Since zero-gap tokens contribute nothing,
$\mathcal L_{\mathrm{opd}}
=
\mathcal L_{\mathrm{opd}}^{+}
+
\mathcal L_{\mathrm{opd}}^{-}$.
Outside the clipped regions, minimizing these components encourages
increasing the likelihood of tokens assigned higher probability by
the teacher and decreasing that of tokens assigned lower probability,
respectively.

\clearpage
\section{Experimental Configurations}
\label{app:exp_set}
In this section, we describe the experimental configurations for NPO, DPO, PPO, GRPO, and OPD, including training, endpoint construction, and performance evaluation. We then detail the shared endpoint-probing procedures and protocols for comparing controlled variants with their baselines.
All experiments were conducted on servers equipped with NVIDIA H100 80GB HBM3 and A100 80GB SXM4 GPUs and AMD EPYC 7V13 CPUs. % Our source code is available at \url{https://anonymous.4open.science/r/Negative-learning-C5A6} and will be fully open-sourced upon acceptance.

\subsection{NPO}
\label{app:setup-npo}

\textbf{Data and initialization.}
Following the NPO setup~\citep{zhang2024negative} and the OpenUnlearning implementation~\citep{dorna2026openunlearning}, we performed unlearning on the TOFU \texttt{forget05} split, which contains $200$ forget examples. We initialized Llama3.2-1B and Llama3.2-3B from the TOFU-full checkpoints provided by OpenUnlearning.

\textbf{Training.}
We used NPO with $\beta=0.1$ without retention regularization, leaving only the negative-learning term. We trained for $10$ epochs with a batch size of $32$, corresponding to $70$ optimizer updates, using AdamW with a learning rate of $10^{-5}$, weight decay of $0.01$, one epoch of warmup followed by linear decay, and gradient clipping at a maximum norm of $1$. The maximum sequence length was $512$ tokens. These settings follow the default NPO configuration provided by OpenUnlearning.

\textbf{Endpoint construction and probing.}
For each model, we constructed $\mathcal D_{\mathrm{end}}$ by pairing the rephrased forget prompts provided by TOFU with responses greedily decoded from the final step-$70$ checkpoint, using a maximum generation length of $128$ tokens. We probed checkpoints every $10$ steps.

\subsection{DPO}
\label{app:setup-dpo}
\textbf{Data and initialization.}
Following~\citet{tunstall2023zephyr}, we first performed supervised fine-tuning on UltraChat-200k for all base models, followed by DPO training on UltraFeedback. We used Pythia-2.8B and Qwen3-1.7B for endpoint probing and additionally evaluated Llama3-8B in the controlled-method comparisons. % Both DPO and PPO, discussed next, were implemented using the TRL library\footnote{\url{https://huggingface.co/docs/trl}}.

\textbf{Training.}
All three models used $\beta=0.1$ and were trained for two epochs with AdamW and weight decay $0.01$. The model-specific configurations were:
\begin{center}
\begin{tabular}{lcc}
\hline
Model & Global batch size & Initial learning rate \\
\hline
Pythia-2.8B & 32 & $5\times10^{-7}$ \\
Qwen3-1.7B & 512 & $5\times10^{-5}$ \\
Llama3-8B & 256 & $1\times10^{-5}$ \\
\hline
\end{tabular}
\end{center}
For Pythia-2.8B, we used a linear learning-rate schedule with a $5\%$ warmup ratio and gradient clipping at a maximum norm of $1$.

\textbf{Endpoint construction and probing.}
For in-distribution probing, we used held-out UltraFeedback prompts. For out-of-distribution probing, we used HH-RLHF-helpfulness prompts. By default, each prompt was paired with a response greedily decoded from the final DPO checkpoint. We evaluated the full DPO objective and its positive- and negative-learning components at the same jointly trained checkpoints.

\textbf{Performance evaluation.}
We evaluated preference alignment on AlpacaEval using GPT-5.5 as the judge in an \emph{LLM-as-a-judge} framework~\citep{dubois2024length}.
Candidate responses were greedily decoded with a maximum generation length of $512$ tokens. 
We compared each candidate model with its corresponding supervised fine-tuning baseline and reported the raw win rate without length correction, measuring how often the judge preferred the candidate response. These evaluations test whether the probing analysis motivates useful interventions.

\subsection{PPO}
\label{app:setup-ppo}

\textbf{Data and initialization.}
As in DPO, we initialized PPO from models supervised fine-tuned on UltraChat-200k and used UltraFeedback prompts for online rollouts. Pythia-2.8B and Qwen3-1.7B were used for probing, with Llama3-8B additionally included in the controlled-method comparisons.

\textbf{Training and rollout generation.}
We used AdamW and gradient clipping at a maximum norm of $1$. We sampled one response per prompt with top-$p=0.9$. The model-specific configurations were:
\begin{center}
\small
\setlength{\tabcolsep}{4pt}
\begin{tabular}{lcccc}
\hline
Model & Global batch & Learning rate & Temperature & Max. new tokens \\
\hline
Pythia-2.8B & 128 & $2\times10^{-6}$ & 1.0 & 256 \\
Qwen3-1.7B & 128 & $2\times10^{-6}$ & 0.7 & 400 \\
Llama3-8B & 32 & $5\times10^{-7}$ & 0.7 & 400 \\
\hline
\end{tabular}
\end{center}
For Qwen3-1.7B and Llama3-8B, the maximum prompt length was 256 tokens. Pythia-2.8B and Qwen3-1.7B rollouts were scored by ArmoRM-Llama3-8B-v0.1~\citep{ArmoRM}, whereas Llama3-8B rollouts were scored by Skywork-Reward-Llama3.1-8B~\citep{liu2024skywork}. For Pythia-2.8B, the reported training trajectory extended to step $2400$, with endpoint probing performed every $400$ steps.

\textbf{Endpoint construction and probing.}
We constructed endpoint datasets from held-out HH-RLHF-helpfulness prompts paired with responses greedily decoded from the final PPO checkpoints. At each probed checkpoint, we computed endpoint-probing scores for the clipped policy surrogate and its positive- and negative-advantage components. 

\textbf{Performance evaluation.}
We used the same AlpacaEval protocol as for DPO, with greedy decoding capped at $512$ generated tokens and comparison against the corresponding supervised fine-tuning baseline.

\subsection{GRPO}
\label{app:setup-grpo}
\label{app:idot_reviewer}
\label{app:iidd_revw}

We evaluated GRPO in three settings: single-turn function calling, multi-turn retail interaction, and mathematical reasoning. % All implementations were based on VERL\footnote{\url{https://verl.readthedocs.io}}.

\textbf{Single-Turn Function Calling.}
We used Gemma4-E4B (the instruction-tuned checkpoint) on Hermes Function-Calling v1. The experiment used approximately $2000$ training prompts and $300$ held-out evaluation prompts. We applied no additional supervised fine-tuning before GRPO.
We trained for $58$ updates, using $32$ prompts per update and a group size of $G=5$, yielding $160$ sampled responses per update. We applied \emph{low-rank adaptation} (LoRA)~\citep{hu2021lora} to all linear layers with rank $32$ and $\alpha=64$. We used AdamW with a learning rate of $10^{-6}$ and a KL coefficient of $10^{-3}$. Responses were sampled at temperature $1.0$ with a maximum generation length of $1024$ tokens and scored by a programmatic function-calling reward.
{Following the shared endpoint-construction protocol, we paired held-out prompts with responses greedily decoded from the final model. } 
We greedily decoded responses to $300$ held-out prompts, scored them using the training reward function, and reported the mean programmatic reward. 

\textbf{Multi-Turn Retail Interaction.}
We initialized Qwen3.5-9B from a retail-specific supervised fine-tuning checkpoint and trained on the retail domain of $\tau^2$-Bench. Customer responses were generated by a frozen Qwen3.5-4B simulator.
Each update used eight prompts and $G=8$. We trained a text-tower LoRA adapter with rank $32$ and $\alpha=64$ using AdamW, a learning rate of $10^{-4}$, and a KL coefficient of $10^{-3}$. Policy responses were sampled at temperature $1.0$, with at most $512$ generated tokens per assistant turn, $16$ assistant turns, and $12$ simulated-user replies. Binary rewards were determined from the final environment state.
We evaluated one fixed-seed trajectory on retail evaluation tasks and reported the task-success rate.

\textbf{Mathematical Reasoning.}
We trained Qwen3-0.6B on MATH problems, balanced between Levels 4 and 5. We applied no additional supervised fine-tuning before GRPO.
Each update used eight prompts and $G=8$. We performed full-parameter training with AdamW, a learning rate of $10^{-6}$, and a KL coefficient of $10^{-3}$. Responses were sampled at temperature $0.6$, top-$p=0.95$, and top-$k=20$, with a maximum generation length about $16000$ tokens. Rewards were assigned by exact-answer verification.
We evaluated checkpoints on MATH500, using one fixed-seed response per problem, and reported pass@1.

\subsection{OPD}
\label{app:setup-opd}

\textbf{Data and initialization.}
We use Qwen3-1.7B-Base as the student and a frozen Qwen3-1.7B as the teacher, and train on \texttt{guanning-ai/dapo14k}~\citep{yu2026dapo}. We adapt the sampled-token OPD baseline released with vOPD~\citep{oh2026kl}.

\textbf{Training.}
We train with LoRA using rank $64$ and scaling factor $128$, a learning rate of $10^{-5}$ with cosine decay, and a global batch size of $64$ prompts.
Training uses distributed data parallelism and runs for one epoch, comprising $211$ updates over $13{,}504$ examples. We conduct one training run with seed $42$.

\textbf{Endpoint construction and probing.}
We constructed two endpoint datasets, each containing $64$ prompts,
drawn from held-out data and examples consumed during training,
respectively. Each prompt was paired with a response sampled from
the final step-$211$ student checkpoint, using temperature $1.0$,
top-$p=1.0$, and a maximum generation length of $2048$ tokens.
At each probed checkpoint, we computed endpoint-probing scores
for the OPD surrogate and its positive- and negative-learning
components over the trainable LoRA parameters.

\subsection{Endpoint-Probing Implementation}
\label{app:setup-probing}

\textbf{Endpoint choice.}
% Endpoint datasets paired held-out prompts with responses greedily decoded from the final checkpoint. Greedy decoding removes response-sampling randomness for a fixed prompt and checkpoint. 
Unless otherwise specified, endpoint datasets paired held-out prompts with responses greedily decoded from the final checkpoint. For these endpoints, greedy decoding removes response-sampling randomness for a fixed prompt and checkpoint.
The probe is retrospective and measures whether intermediate updates support the realized final responses, not whether these responses are optimal. We compare component effects under matched endpoints and assess task performance separately using the metrics specified above.

\textbf{Gradient estimation.}
Component gradients were evaluated at the same jointly trained model state, using the decompositions in Appendix~\ref{app:po method}. To reduce computation in the DPO and PPO analyses, we sampled $500$ examples from the objective data and $500$ examples from $\mathcal D_{\mathrm{end}}$. We partitioned each sample into mini-batches of four examples for DPO and six for PPO to limit GPU memory usage. We averaged the endpoint gradients across mini-batches before computing the probing scores. % We length-normalized log-likelihoods to reduce sensitivity to response length when computing $\delta$.

\textbf{Approximation validation.}
To validate the first-order approximation in Figure~\ref{fig:correlation}, we compared the endpoint-probing score $\delta$ with the exact one-step endpoint-loss reduction $\Delta$ for DPO on Qwen3-1.7B and PPO on Pythia-2.8B. At each probed checkpoint, we computed $\delta$ and applied one gradient update to a copy of the model to measure $\Delta$, using the same objective mini-batch and endpoint data for both quantities. Endpoint prompts were drawn from HH-RLHF-helpfulness, with responses greedily decoded from the final checkpoint. We kept the original checkpoint unchanged across comparisons.

% \textbf{Gradient-norm filtering.} We used interquartile-range (IQR) filtering to reduce the influence of mini-batches with extreme gradient norms, echoing the typical use of gradient clipping in LLM fine-tuning. Given the mini-batch gradient norms, we computed the first and third quartiles, $Q_1$ and $Q_3$, and excluded batches exceeding $Q_3+1.5(Q_3-Q_1)$ from the probing estimate. 

% \textbf{Gradient-norm filtering.} Since probing is performed post hoc, we use interquartile-range (IQR) filtering to approximate the influence-limiting effect of gradient clipping commonly used in LLM fine-tuning. Given the mini-batch gradient norms, we compute the first and third quartiles, $Q_1$ and $Q_3$, and exclude batches exceeding $Q_3+1.5(Q_3-Q_1)$ from the probing estimate. Although filtering removes rather than rescales extreme gradients, both operations prevent a small number of high-norm batches from dominating the estimate~\citep{pascanu2013difficulty,merad2023robust}.

\subsection{DPO Component Ablations}
\label{app:dpo-ablation}

We detail the ablations in Figure~\ref{fig:dpo ablation}, which examine whether components with negative endpoint-probing scores remain beneficial to fine-tuning performance.

\textbf{Removing negative learning early.}
We initialized from the same supervised fine-tuning checkpoints as standard DPO and trained with only $\mathcal L_{\mathrm{dpo}}^{+}$.
We evaluated checkpoints through step $3000$ for Pythia-2.8B and step $120$ for Qwen3-1.7B, corresponding to the early phase when negative learning opposes endpoint fitting. These results are labeled \emph{w/o NEG}.

\textbf{Removing positive learning later.}
We initialized from the standard DPO checkpoints at step $6000$ for Pythia-2.8B and step $180$ for Qwen3-1.7B, when positive learning opposes endpoint fitting. We then continued training with only $\mathcal L_{\mathrm{dpo}}^{-}$. These results are labeled \emph{w/o POS}.
Training steps retain the numbering of the original DPO runs.

For both ablations, we removed the specified component without
rescaling the remaining component. All other training settings and the AlpacaEval evaluation protocol remained the same as for standard DPO.

\subsection{Controlled Negative Learning}
\label{app:setup-controlled}

We detail the scheduling rules used to coordinate positive and negative learning in cDPO, cPPO, and cGRPO, followed by the hyperparameters selected on held-out validation data.

\textbf{cDPO.}
We gradually shift the relative emphasis from positive to negative learning to reflect their changing roles during training. Our DPO analysis shows that positive learning drives target shaping early, whereas negative learning becomes endpoint-aligned later for the out-of-distribution endpoints studied. We therefore use
\begin{equation}
\mathcal L_{\mathrm{cdpo}}(\boldsymbol\theta)
=
2(1-\lambda_t)\mathcal L_{\mathrm{dpo}}^{+}(\boldsymbol\theta)
+
2\lambda_t\mathcal L_{\mathrm{dpo}}^{-}(\boldsymbol\theta),
\qquad
\lambda_t=\max\big(\min\big(\frac{t-t_1}{t_2-t_1},1\big),0\big).
\end{equation}
Here, $t_1<t_2$ specify the transition interval, during which $\lambda_t$ increases linearly from $0$ to $1$. For the special case $t_1=t_2$, we use an abrupt transition, with $\lambda_t=0$ before the transition step and $\lambda_t=1$ from that step onward.
The factors of two recover standard DPO at $\lambda_t=1/2$. The transition interval may extend beyond the training horizon $[0,T]$: setting $t_1<0$ and $t_2>T$ keeps both components active throughout training.

\textbf{cPPO and cGRPO.}
In the PPO and GRPO settings studied, negative learning counteracts endpoint-aligned positive learning.
We control the strength of negative learning to examine whether coordinating the two components can improve fine-tuning performance. The controlled policy surrogates are
\begin{equation}
\begin{aligned}
\mathcal L_{\mathrm{cppo}}(\boldsymbol\theta)
&=
\mathcal L_{\mathrm{ppo}}^{+}(\boldsymbol\theta)
+\lambda_t\mathcal L_{\mathrm{ppo}}^{-}(\boldsymbol\theta),\\
\mathcal L_{\mathrm{cgrpo}}(\boldsymbol\theta)
&=
\mathcal L_{\mathrm{grpo}}^{+}(\boldsymbol\theta)
+\lambda_t\mathcal L_{\mathrm{grpo}}^{-}(\boldsymbol\theta).
\end{aligned}
\end{equation}
Here, $\lambda_t\geq0$ controls the strength of negative learning, with $\lambda_t=1$ recovering the standard policy surrogate and $\lambda_t<1$ and $\lambda_t>1$ attenuating and strengthening
negative learning, respectively. Other objective terms remain unchanged.
We consider three scheduling rules:
\begin{itemize}[leftmargin=1.2em, itemsep=2pt, topsep=0pt]
    \item \textbf{Periodic attenuation.} We periodically weaken negative learning, allowing positive updates to more strongly reinforce currently high-reward responses. We set
    \begin{equation}
    \lambda_t=1+\max\big(\min\big(\sin(\pi t/t_3),0\big),-\tau\big).
    \end{equation}
    Here, $t_3>0$ determines the period $2t_3$, and $\tau\in(0,1]$ controls the maximum attenuation.

    \item \textbf{Constant scaling.} We maintain negative learning at a fixed reduced or enhanced strength, aiming to favor exploitation or exploration, respectively. We set $\lambda_t=c$, where $c\geq0$. Here, $c<1$ reduces the strength of negative learning, whereas $c>1$ enhances it.

    \item \textbf{Adaptive strengthening.} For GRPO, we increase the strength of negative learning as the fraction of fully successful rollout groups rises, aiming to encourage further exploration. We set $\lambda_t=1+\kappa s_t$, where $\kappa\geq0$. Here, $s_t$ is the fraction of rollout groups in the batch in which all responses are successful, and $\kappa$ controls the maximum increase in negative-learning strength.
\end{itemize}

\textbf{Hyperparameter selection.}
We tuned the scheduling hyperparameters on held-out validation data and selected the configurations below for the main comparisons. In the table, Periodic, Constant, and Adaptive refer to periodic attenuation, constant scaling, and adaptive strengthening, respectively.

\begin{center}
\small
\setlength{\tabcolsep}{4pt}
\begin{tabular}{ccc}
\hline
Method & Model & Control hyperparameters \\
\hline
cDPO  & Pythia-2.8B   & $t_1=2500,\ t_2=6500$ \\
cDPO  & Qwen3-1.7B    & $t_1=100,\ t_2=200$ \\
cDPO  & Llama3-8B     & $t_1=20,\ t_2=60$ \\
\hline
cPPO  & Pythia-2.8B   & Periodic: $t_3=2,\ \tau=0.08$ \\
cPPO  & Qwen3-1.7B    & Periodic: $t_3=100,\ \tau=0.01$ \\
cPPO  & Llama3-8B     & Periodic: $t_3=5,\ \tau=0.99$ \\
\hline
cGRPO & Gemma4-E4B    & Periodic: $t_3=5,\ \tau=0.001$ \\
cGRPO & Qwen3.5-9B    & Adaptive: $\kappa=0.8$ \\
cGRPO & Qwen3-0.6B    & Constant: $c=0.9$ \\
\hline
\end{tabular}
\end{center}

\clearpage
\section{Additional Experiments}
\label{app:support}

This section provides additional probing results and presents further controlled-method comparisons.

\subsection{Endpoint Probing on Main Settings}
\label{app:complete-probing}

Figure~\ref{fig:ppo-normalized} complements the endpoint-probing scores with cosine similarities and gradient norms, based on the decomposition
$\delta(\mathcal L;\boldsymbol\theta)=\|\nabla\mathcal L_{\mathrm{end}}(\boldsymbol\theta)\|\|\nabla\mathcal L(\boldsymbol\theta)\|\cos\!\left(\nabla\mathcal L_{\mathrm{end}}(\boldsymbol\theta),\nabla\mathcal L(\boldsymbol\theta)\right)$.
These quantities distinguish gradient direction from scale, as a small probing score may reflect either small gradient norms or weak directional alignment. We extend this analysis to the experiments below. All quantities within each experiment were evaluated at the same checkpoints using the same endpoint set.

\begin{figure*}[t]
    \centering
    \begin{subfigure}[t]{0.3\textwidth}
        \centering
        \includegraphics[width=\textwidth]{new_fig/npo_llama3b.pdf}
        \caption{{Endpoint-probing score}}\label{fig:app npo 3b score}
    \end{subfigure}\quad
    \begin{subfigure}[t]{0.3\textwidth}
        \centering
        \includegraphics[width=\textwidth]{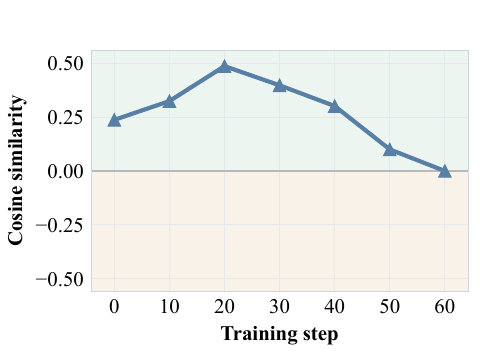}
        \caption{{Cosine similarity}}\label{fig:app npo 3b cos}
    \end{subfigure}\quad
    \begin{subfigure}[t]{0.3\textwidth}
        \centering
        \includegraphics[width=\textwidth]{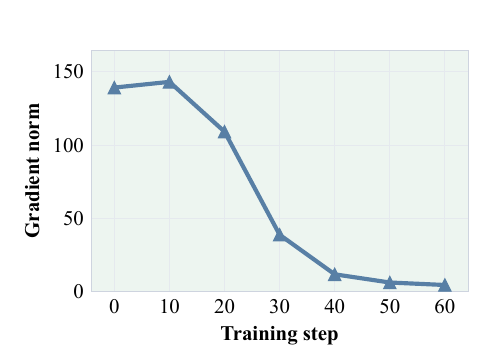}
        \caption{{Gradient norm}}\label{fig:app npo 3b norm}
    \end{subfigure}
    \caption{\textbf{NPO Probing on Llama3.2-3B.} We report endpoint-probing scores, cosine similarities between the NPO and endpoint gradients, and their gradient norms. }
\label{fig:app npo 3b}
\end{figure*}

\textbf{NPO.}
Figure~\ref{fig:app npo 3b} provides the complete NPO results on Llama3.2-3B. The positive probing scores across the evaluated checkpoints confirm that negative learning exhibits target shaping within this setting, consistent with previous findings~\citep{zhang2024negative,zhu2026surprising}. However, the cosine similarity gradually approaches zero, indicating that endpoint alignment weakens during training. This trend raises the possibility that target shaping may not persist under longer training, echoing previous reports of instability in negative-only learning~\citep{wang2025rethinking,li2026llm}.

\begin{figure*}[t]
    \centering
    \begin{subfigure}[t]{0.3\textwidth}
        \centering
        \includegraphics[width=\textwidth]{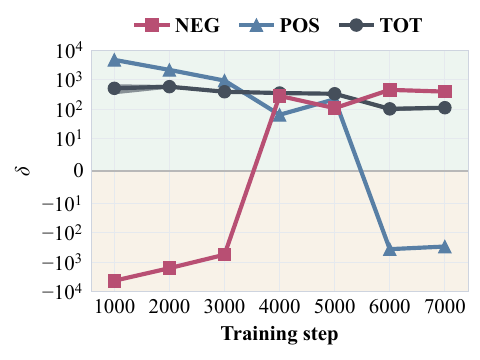}
        \caption{Endpoint-probing score}
        \label{fig:app dpo pythia score}
    \end{subfigure}\quad
    \begin{subfigure}[t]{0.3\textwidth}
        \centering
        \includegraphics[width=\textwidth]{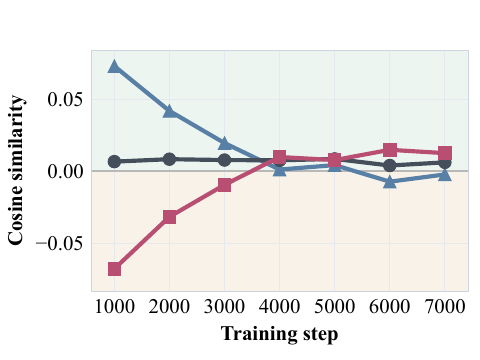}
        \caption{{Cosine similarity}}
        \label{fig:app dpo pythia cos}
    \end{subfigure}\quad
    \begin{subfigure}[t]{0.3\textwidth}
        \centering
        \includegraphics[width=\textwidth]{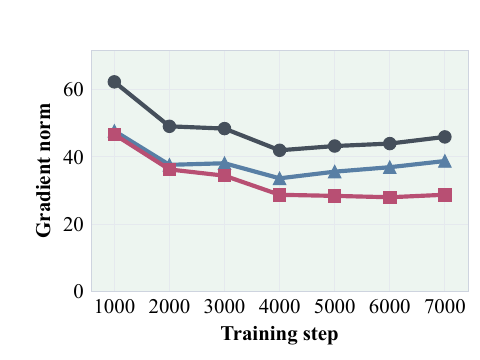}
        \caption{Gradient norm}
        \label{fig:app dpo pythia norm}
    \end{subfigure}
    \caption{\textbf{DPO Probing on Pythia-2.8B.} We report endpoint-probing scores, cosine similarities with the endpoint gradient, and gradient norms for the full DPO objective and its positive and negative components. }
    \label{fig:app dpo pythia}
\end{figure*}

\begin{figure*}[t]
    \centering
    \begin{subfigure}[t]{0.3\textwidth}
        \centering
        \includegraphics[width=\textwidth]{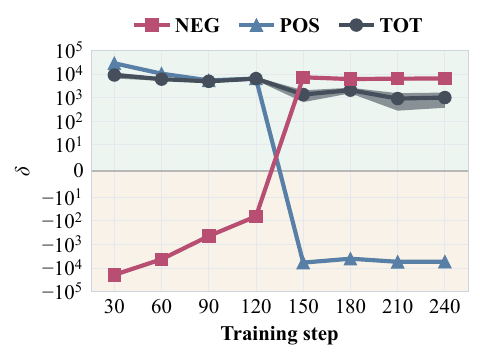}
        \caption{Endpoint-probing score}
        \label{fig:app dpo qwen score}
    \end{subfigure}\quad
    \begin{subfigure}[t]{0.3\textwidth}
        \centering
        \includegraphics[width=\textwidth]{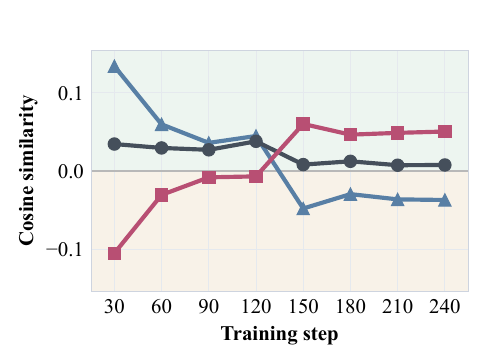}
        \caption{{Cosine similarity}}
        \label{fig:app dpo qwen cos}
    \end{subfigure}\quad
    \begin{subfigure}[t]{0.3\textwidth}
        \centering
        \includegraphics[width=\textwidth]{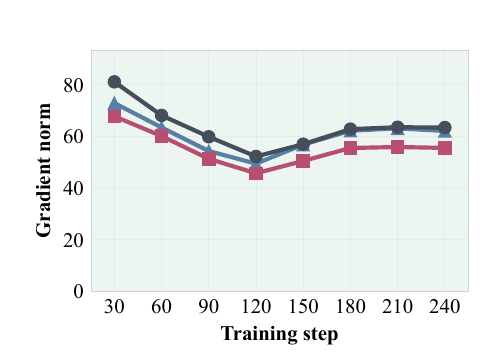}
        \caption{Gradient norm}
        \label{fig:app dpo qwen norm}
    \end{subfigure}
    \caption{\textbf{DPO Probing on Qwen3-1.7B.} We report endpoint-probing scores, cosine similarities with the endpoint gradient, and gradient norms for the full DPO objective and its positive and negative components. }
    \label{fig:app dpo qwen}
\end{figure*}

\begin{figure*}[t]
    \centering
    \begin{subfigure}[t]{0.3\textwidth}
        \centering
        \includegraphics[width=\textwidth]{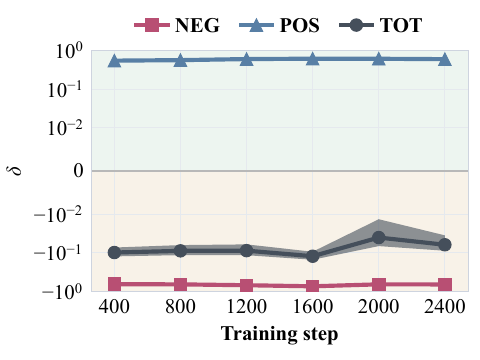}
        \caption{Endpoint-probing score}
        \label{fig:app ppo pythia score}
    \end{subfigure}\quad
    \begin{subfigure}[t]{0.3\textwidth}
        \centering
        \includegraphics[width=\textwidth]{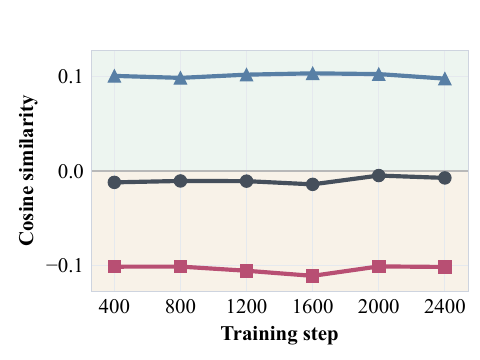}
        \caption{Cosine similarity}
        \label{fig:app ppo pythia cos}
    \end{subfigure}\quad
    \begin{subfigure}[t]{0.3\textwidth}
        \centering
        \includegraphics[width=\textwidth]{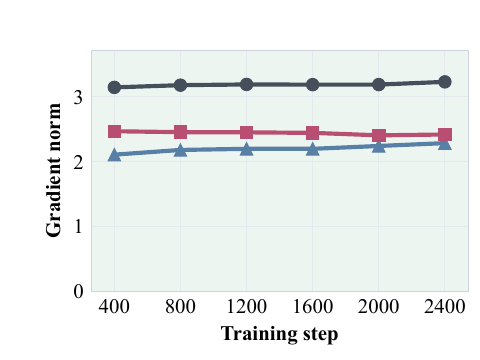}
        \caption{{Gradient norm}}
        \label{fig:app ppo pythia norm}
    \end{subfigure}
    \caption{\textbf{PPO Probing on Pythia-2.8B.} We report endpoint-probing scores, cosine similarities with the endpoint gradient, and gradient norms for the full PPO policy surrogate and its positive and negative components. }
    \label{fig:app ppo pythia}
\end{figure*}

\begin{figure*}[t]
    \centering
    \begin{subfigure}[t]{0.3\textwidth}
        \centering
        \includegraphics[width=\textwidth]{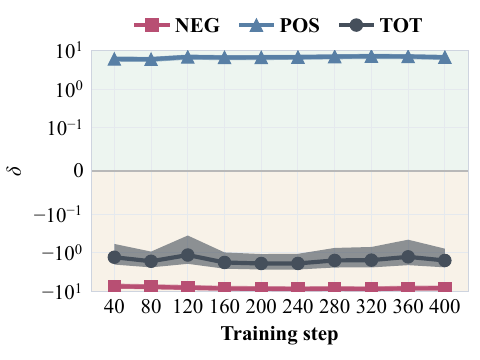}
        \caption{Endpoint-probing score}
        \label{fig:app ppo qwen score}
    \end{subfigure}\quad
    \begin{subfigure}[t]{0.3\textwidth}
        \centering
        \includegraphics[width=\textwidth]{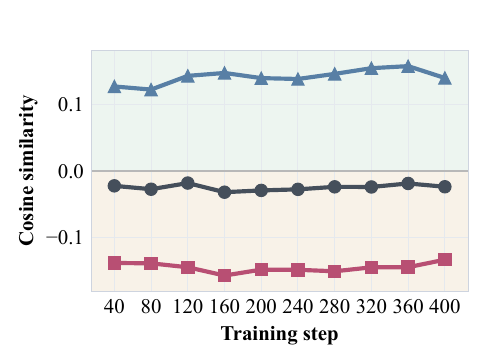}
        \caption{{Cosine similarity}}
        \label{fig:app ppo qwen cos}
    \end{subfigure}\quad
    \begin{subfigure}[t]{0.3\textwidth}
        \centering
        \includegraphics[width=\textwidth]{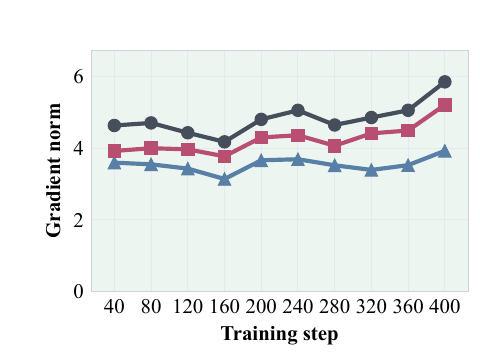}
        \caption{Gradient norm}
        \label{fig:app ppo qwen norm}
    \end{subfigure}
    \caption{\textbf{PPO Probing on Qwen3-1.7B.} We report endpoint-probing scores, cosine similarities with the endpoint gradient, and gradient norms for the full PPO policy surrogate and its positive and negative components. }
    \label{fig:app ppo qwen}
\end{figure*}

\textbf{DPO and PPO.}
Figures~\ref{fig:app dpo pythia}--\ref{fig:app ppo qwen} report the complete results for Pythia-2.8B and Qwen3-1.7B. For DPO, the cosine similarities show the same phase-dependent transition as the probing scores. Positive learning is aligned with the endpoints early, while negative learning becomes aligned later. The gradient norms provide complementary information but do not alone explain this transition. For PPO, positive and negative learning remain oppositely aligned with the endpoints across checkpoints. Their contributions largely cancel, leaving the full-surrogate score small despite non-negligible component gradients. Thus, the small overall score reflects opposing contributions from positive and negative learning rather than weak gradients.

\begin{figure*}[t]
    \centering
    \begin{subfigure}[t]{0.3\textwidth}
        \centering
        \includegraphics[width=\textwidth]{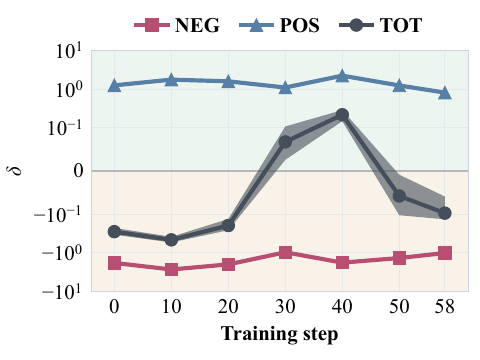}
        \caption{{{Endpoint-probing score}}}
        \label{fig:app grpo fc score}
    \end{subfigure}\quad
    \begin{subfigure}[t]{0.3\textwidth}
        \centering
        \includegraphics[width=\textwidth]{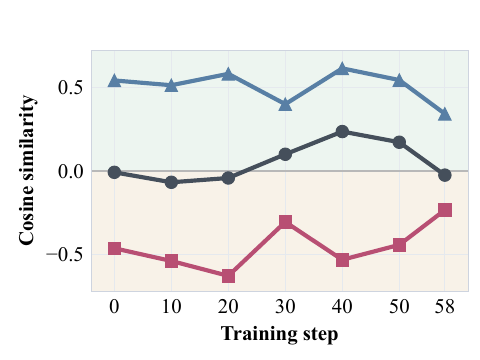}
        \caption{{Cosine similarity}}
        \label{fig:app grpo fc cos}
    \end{subfigure}\quad
    \begin{subfigure}[t]{0.3\textwidth}
        \centering
        \includegraphics[width=\textwidth]{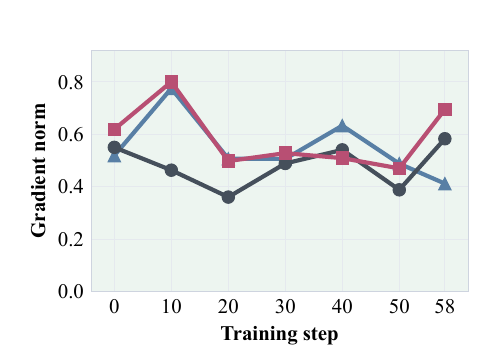}
        \caption{{Gradient norm}}
        \label{fig:app grpo fc norm}
    \end{subfigure}
    \caption{\textbf{GRPO Probing on Hermes Function-Calling v1.} We report endpoint-probing scores for the full GRPO policy surrogate and its positive and negative components, together with component cosine similarities with the endpoint gradient and gradient norms. }
    \label{fig:app grpo fc}
\end{figure*}

\begin{figure*}[t]
    \centering
    \begin{subfigure}[t]{0.3\textwidth}
        \centering
        \includegraphics[width=\textwidth]{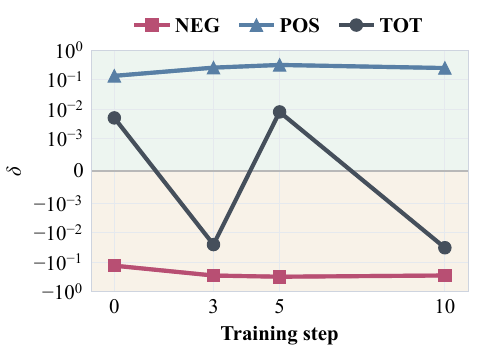}
        \caption{{Endpoint-probing score}}
        \label{fig:app grpo agent score}
    \end{subfigure}\quad
    \begin{subfigure}[t]{0.3\textwidth}
        \centering
        \includegraphics[width=\textwidth]{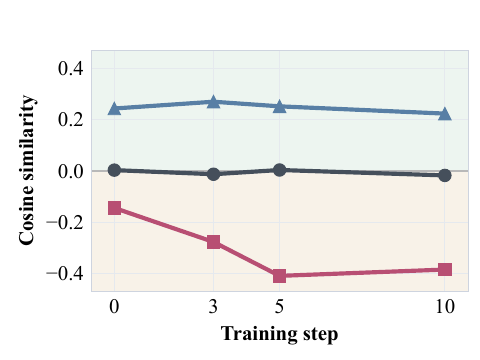}
        \caption{{Cosine similarity}}
        \label{fig:app grpo agent cos}
    \end{subfigure}\quad
    \begin{subfigure}[t]{0.3\textwidth}
        \centering
        \includegraphics[width=\textwidth]{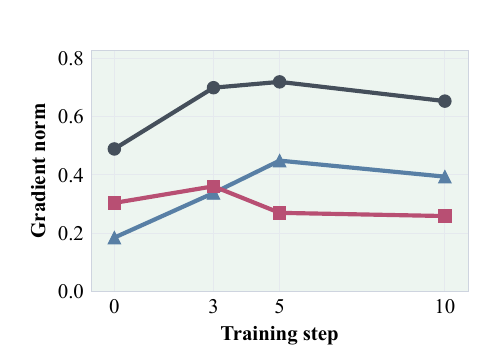}
        \caption{{Gradient norm}}
        \label{fig:app grpo agent norm}
    \end{subfigure}
    \caption{\textbf{GRPO Probing on $\tau^2$ Retail.} We report endpoint-probing scores for the full GRPO policy surrogate and its positive and negative components, together with component cosine similarities with the endpoint gradient and gradient norms. We used Qwen3.5-9B for multi-turn tool use in the $\tau^2$ Retail environment.}
    \label{fig:app grpo agent}
\end{figure*}

\begin{figure*}[t]
    \centering
    \begin{subfigure}[t]{0.3\textwidth}
        \centering
        \includegraphics[width=\textwidth]{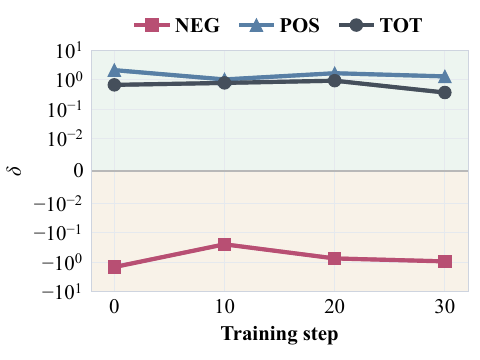}
        \caption{Endpoint-probing score}
        \label{fig:app grpo math score}
    \end{subfigure}\quad
    \begin{subfigure}[t]{0.3\textwidth}
        \centering
        \includegraphics[width=\textwidth]{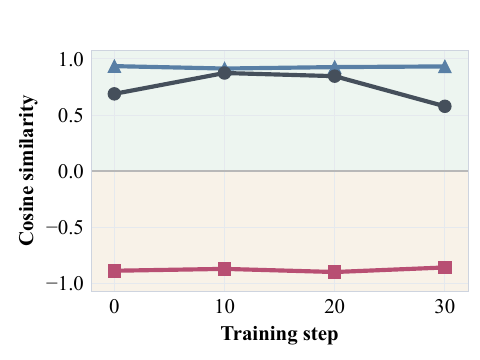}
        \caption{{Cosine similarity}}
        \label{fig:app grpo math cos}
    \end{subfigure}\quad
    \begin{subfigure}[t]{0.3\textwidth}
        \centering
        \includegraphics[width=\textwidth]{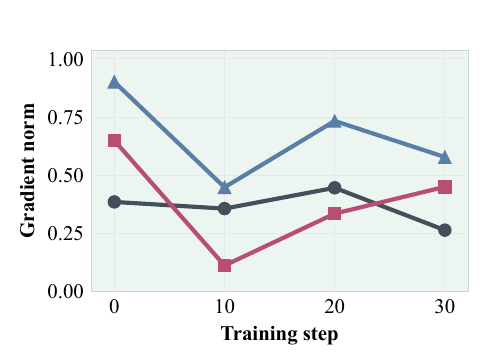}
        \caption{{Gradient norm}}
        \label{fig:app grpo math norm}
    \end{subfigure}
    \caption{\textbf{GRPO Probing on MATH.} We report endpoint-probing scores for the full GRPO policy surrogate and its positive and negative components, together with component cosine similarities with the endpoint gradient and gradient norms. We used Qwen3-0.6B for mathematical reasoning on MATH.}
    \label{fig:app grpo math}
\end{figure*}

\textbf{GRPO.}
The main text examines GRPO on function calling using a single model. We further vary both the model and task to assess whether the observed behavior extends to multi-turn tool use and mathematical reasoning.
We probed Gemma4-E4B on Hermes Function-Calling v1 at checkpoints $\{0,10,20,30,40,50,58\}$, Qwen3.5-9B on $\tau^2$ Retail at $\{0,3,5,10\}$, and Qwen3-0.6B on MATH at $\{0,10,20,30\}$. Their endpoint sets contained $128$, $16$, and $64$ final-policy responses or trajectories, respectively, and remained fixed across checkpoints. Figures~\ref{fig:app grpo fc}--\ref{fig:app grpo math} report the probing scores, cosine similarities, and gradient norms.

Across all settings, positive learning remains directed toward the endpoints, while negative learning moves in the opposite direction, consistent with our observations in Section~\ref{sec:opro}. Their cosine similarities show the same directional separation. 
The full-surrogate score fluctuates across checkpoints, although its magnitude remains small relative to the component scores. On MATH, it even remains positive throughout. This does not contradict our observations about negative learning, which consistently opposes endpoint fitting across all three settings. The sign of the total score instead depends on the relative strengths of the positive and negative contributions, which can be further controlled.

\begin{figure*}[t]
\centering
\begin{subfigure}[t]{\textwidth}
    \centering
    \includegraphics[width=0.32\textwidth]{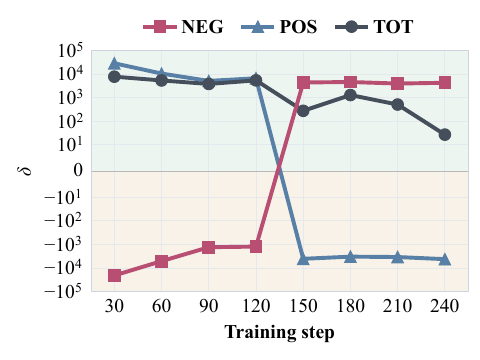}\hfill
    \includegraphics[width=0.32\textwidth]{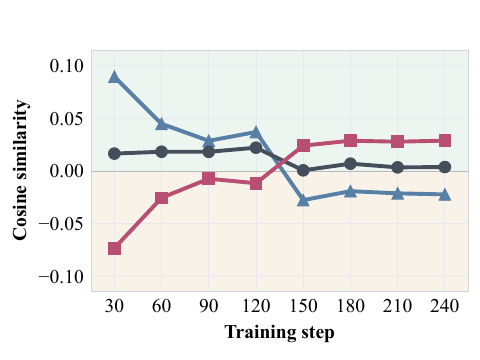}\hfill
    \includegraphics[width=0.32\textwidth]{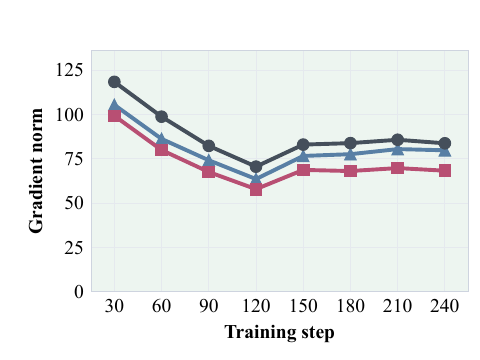}
    \caption{{temperature $0.2$}}
\end{subfigure}

\medskip

\begin{subfigure}[t]{\textwidth}
    \centering
    \includegraphics[width=0.32\textwidth]{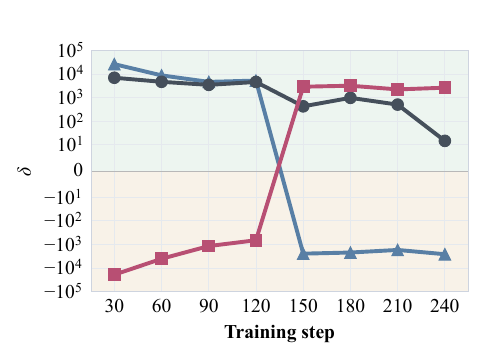}\hfill
    \includegraphics[width=0.32\textwidth]{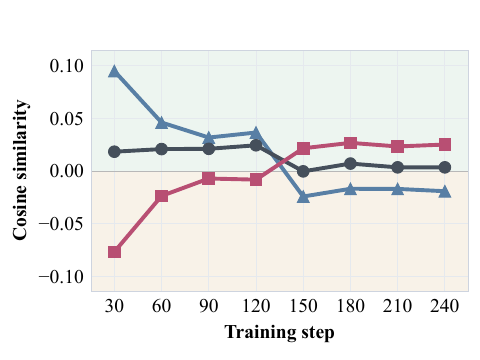}\hfill
    \includegraphics[width=0.32\textwidth]{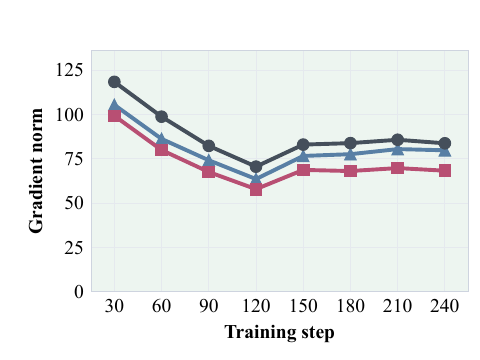}
    \caption{{temperature $0.7$}}
\end{subfigure}

\caption{\textbf{DPO Probing with Sampled Endpoints.}
We report endpoint-probing scores, cosine similarities with the endpoint gradient, and gradient norms for Qwen3-1.7B using responses randomly sampled from the same final checkpoint at the indicated temperatures, rather than greedily decoded responses.}
\label{fig:dpo-endpoint-sampling}
\end{figure*}

\textbf{Sensitivity to sampling.}
To test whether our findings depend on greedy endpoint decoding, we use the same DPO training trajectory as in Figure~\ref{fig:app dpo qwen}, but randomly sample one response per prompt from its final checkpoint at temperatures $0.2$ and $0.7$, both with top-$p=1.0$. Each endpoint set remains fixed across the eight probed checkpoints, with all other probing settings unchanged. Figure~\ref{fig:dpo-endpoint-sampling} shows that both sampling conditions preserve the POS and NEG alignment signs observed with greedy endpoints, including their reversal between steps 120 and 150. Although the magnitudes change, the main pattern persists, supporting the robustness of our findings to various decoding rules.

\subsection{Endpoint Probing on Online DPO}
\label{sec:online-dpo}

We also examine on-policy DPO, which collects fresh preference pairs during training rather than using a fixed dataset as in off-policy DPO. The policy and frozen reference are initialized from the same Qwen3-1.7B checkpoint used in our off-policy experiment. Before each update, we sample two responses for each of $128$ prompts from the current policy and rank them with a frozen Skywork-Reward-V2-Llama-3.1-8B reward model. We train for $240$ updates with $\beta=0.1$ and a peak learning rate of $5\times10^{-7}$. The final checkpoint achieves a raw win rate of $62.11\%$ against the corresponding supervised fine-tuning baseline on $805$ AlpacaEval instructions. For probing, we pair $500$ HH-RLHF-helpfulness prompts with greedy responses from the final checkpoint, capped at $400$ tokens. We estimate objective gradients using a fixed set of $500$ preference pairs sampled from the completed training run. At steps $\{30,60,90,120,150,180,210,240\}$, we recompute the component gradients on these same pairs.

\begin{figure*}[t]
\centering
\begin{minipage}[t]{0.32\textwidth}
    \centering
    \includegraphics[width=\linewidth]{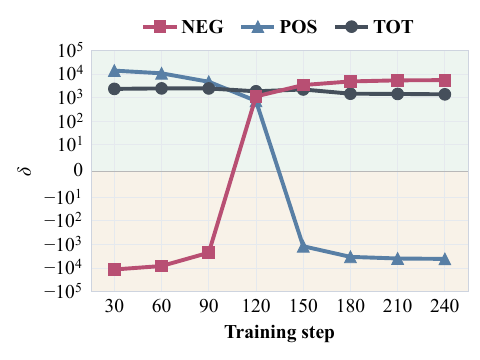}
    \par\smallskip
    {\small (a) Endpoint-probing score}
\end{minipage}\hfill
\begin{minipage}[t]{0.32\textwidth}
    \centering
    \includegraphics[width=\linewidth]{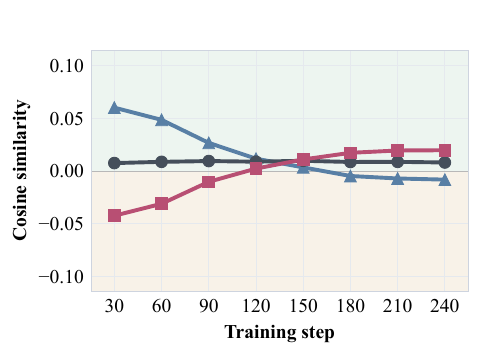}
    \par\smallskip
    {\small (b) Cosine similarity}
\end{minipage}\hfill
\begin{minipage}[t]{0.32\textwidth}
    \centering
    \includegraphics[width=\linewidth]{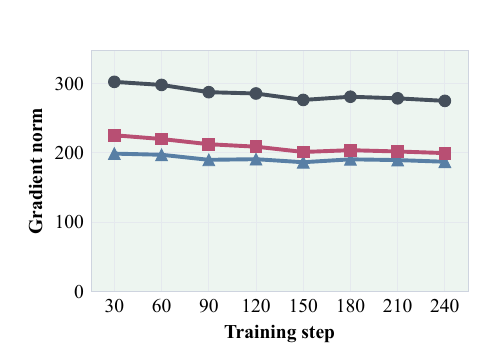}
    \par\smallskip
    {\small (c) Gradient norm}
\end{minipage}
\caption{\textbf{Online DPO Probing on Qwen3-1.7B.}
We report endpoint-probing scores, cosine similarities with the endpoint gradient, and gradient norms for the DPO objective and its positive and negative components. Probing uses fixed recorded preference pairs and greedy final-policy endpoints. Results are from one training run.}
\label{fig:online-dpo-probing}
\end{figure*}

Figure~\ref{fig:online-dpo-probing} shows a transition similar to off-policy DPO. Positive learning initially aligns with the final endpoints, while negative learning opposes them. The probing score for negative learning becomes positive by step $120$, and that for positive learning becomes negative at step $150$. The score for the full DPO objective remains positive but small relative to the component scores. These results suggest that on- versus off-policy data collection alone does not determine component alignment. The objective and its supervision signal may be more relevant to the observed pattern.

\subsection{Endpoint Probing on OPD}
\label{app:opd-probing}

We further examine whether the opposing component alignments observed in PPO and GRPO also arise in other on-policy settings. OPD provides a complementary case: it also trains on student-generated responses, but uses teacher-provided supervision rather than task rewards. We probe Qwen3-1.7B-Base trained with a frozen Qwen3-1.7B teacher, using the training configuration described in Appendix~\ref{app:exp_set}. We consider two endpoint settings: $64$ held-out prompts with a distinct set of $64$ prompts for objective-gradient estimation, and $64$ training prompts shared between the two. The training-prompt setting extends the held-out endpoint construction used in our main analysis. The latter prompts are drawn from examples consumed during the full training run and need not have been encountered by every checkpoint. Thus, this comparison varies both training exposure and prompt overlap. We probe checkpoints at steps $\{0,10,30,50,70,90,110,150,180,210\}$, using one sampled response per prompt for objective-gradient estimation.

\begin{figure*}[t]
    \centering
    \begin{subfigure}[t]{0.32\textwidth}
        \centering
        \includegraphics[width=\textwidth]{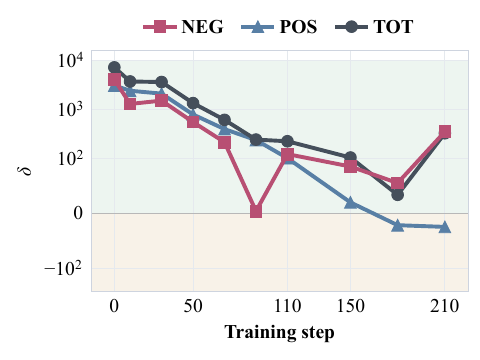}
        \caption{Endpoint-probing score}
    \end{subfigure}\hfill
    \begin{subfigure}[t]{0.32\textwidth}
        \centering
        \includegraphics[width=\textwidth]{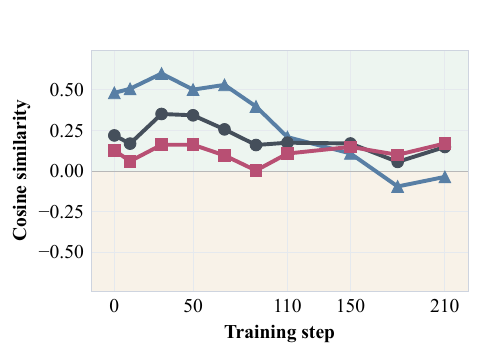}
        \caption{Cosine similarity}
    \end{subfigure}\hfill
    \begin{subfigure}[t]{0.32\textwidth}
        \centering
        \includegraphics[width=\textwidth]{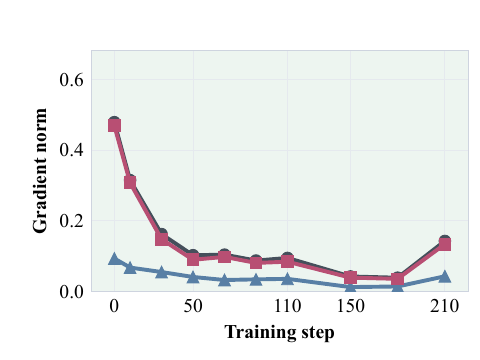}
        \caption{Gradient norm}
    \end{subfigure}

    \medskip

    \begin{subfigure}[t]{0.32\textwidth}
        \centering
        \includegraphics[width=\textwidth]{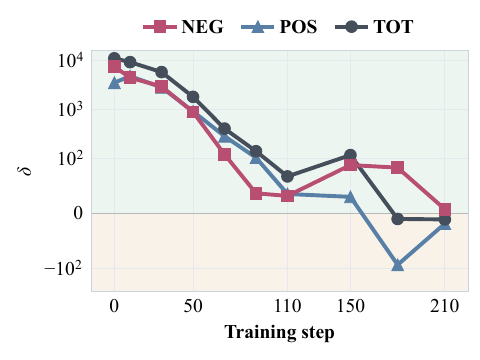}
        \caption{Endpoint-probing score}
    \end{subfigure}\hfill
    \begin{subfigure}[t]{0.32\textwidth}
        \centering
        \includegraphics[width=\textwidth]{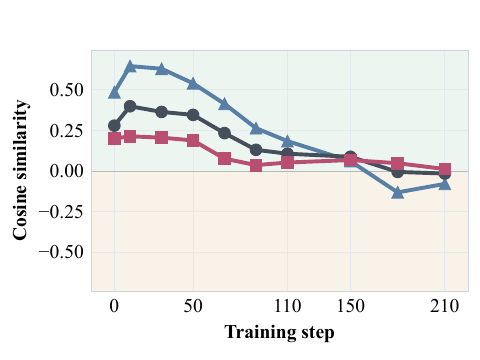}
        \caption{Cosine similarity}
    \end{subfigure}\hfill
    \begin{subfigure}[t]{0.32\textwidth}
        \centering
        \includegraphics[width=\textwidth]{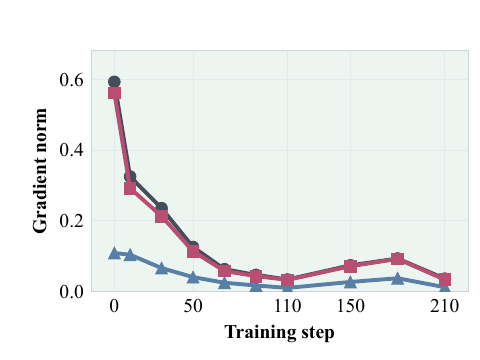}
        \caption{Gradient norm}
    \end{subfigure}

    \caption{\textbf{OPD Probing on Qwen3-1.7B-Base.}
    We report endpoint-probing scores, cosine similarities with
    the endpoint gradient, and gradient norms for the full OPD
    surrogate and its positive and negative components.
    The top row uses distinct held-out prompts for endpoint
    construction and objective-gradient estimation, while the
    bottom row uses the same training prompts for both.
    All results are from one training run.}
    \label{fig:opd-probing}
\end{figure*}

Figure~\ref{fig:opd-probing} reports probing scores, cosine similarities, and gradient norms. In both settings, negative learning remains positively endpoint-aligned across the evaluated checkpoints. Positive learning is initially aligned with the endpoints but becomes negatively aligned at steps $180$ and $210$. The cosine similarities exhibit the same directional change. The full-surrogate score remains positive on held-out prompts but becomes slightly negative at the final two checkpoints in the training-prompt setting. These dynamics differ from the stable opposing component alignments observed in PPO and GRPO.
A possible explanation is that teacher supervision gives negative learning a more consistent target-shaping direction, while positive updates need not remain aligned with the final responses. This echoes our earlier argument that component roles depend on the objective and training phase, suggesting that on-policy sampling alone does not imply the PPO/GRPO pattern.

\subsection{Controlled Methods}
\label{app:coordinate}
\begin{figure*}[t]
    \centering
    \begin{subfigure}[t]{0.49\textwidth}
        \centering
        \includegraphics[width=\textwidth]{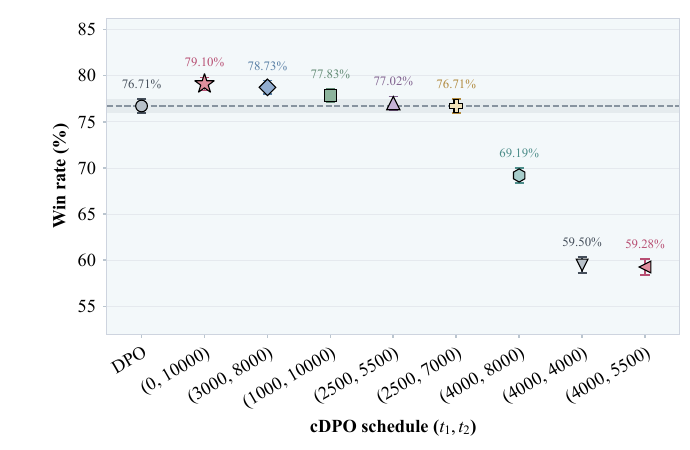}
        \caption{cDPO schedules $(t_1,t_2)$}
        \label{fig:schedules pythia cdpo}
    \end{subfigure}\hfill
    \begin{subfigure}[t]{0.49\textwidth}
        \centering
        \includegraphics[width=\textwidth]{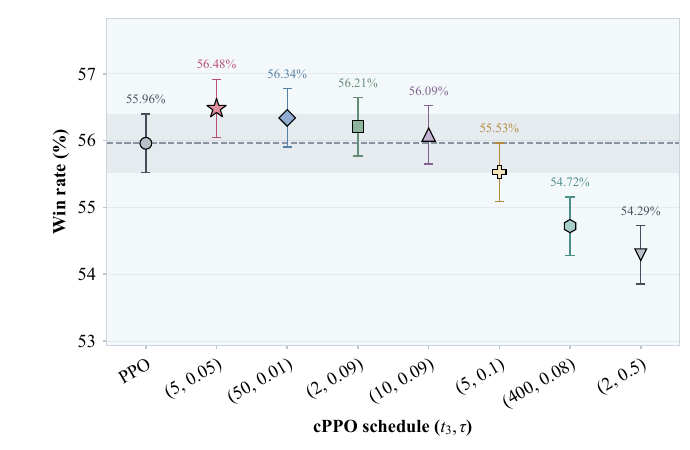}
        \caption{cPPO schedules $(t_3,\tau)$}
        \label{fig:schedules pythia cppo}
    \end{subfigure}
    \caption{\textbf{Controlled Variants on Pythia-2.8B.}
    We compare AlpacaEval win rates across different cDPO and cPPO
    schedules. Markers indicate mean test win rates across three
    independent training seeds, and error bars show $\pm$ one
    standard deviation across seeds. Dashed horizontal lines
    indicate the corresponding DPO and PPO baseline means,
    with shaded bands showing $\pm$ one standard deviation.}
    \label{fig:schedules pythia}
\end{figure*}

\begin{figure*}[t]
    \centering
    \begin{subfigure}[t]{0.49\textwidth}
        \centering
        \includegraphics[width=\textwidth]{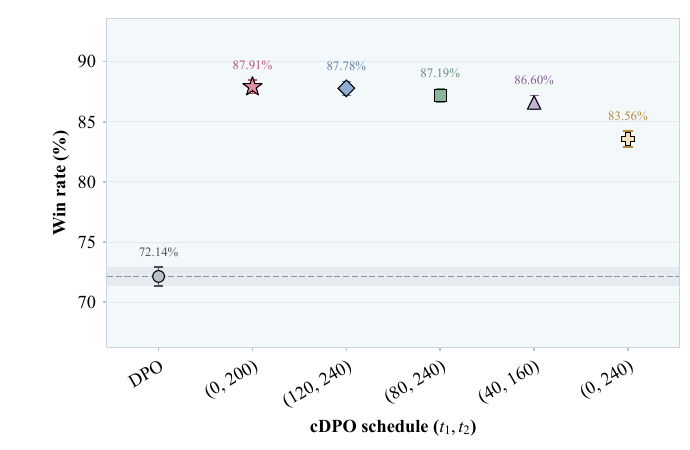}
        \caption{cDPO schedules $(t_1,t_2)$}
        \label{fig:schedules qwen cdpo}
    \end{subfigure}\hfill
    \begin{subfigure}[t]{0.49\textwidth}
        \centering
        \includegraphics[width=\textwidth]{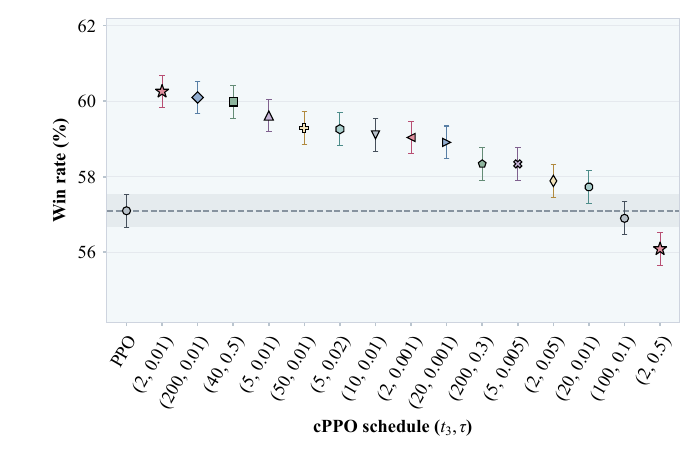}
        \caption{cPPO schedules $(t_3,\tau)$}
        \label{fig:schedules qwen cppo}
    \end{subfigure}
    \caption{\textbf{Controlled Variants on Qwen3-1.7B.}
    We compare AlpacaEval win rates across different cDPO and cPPO
    schedules. Markers indicate mean test win rates across three
    independent training seeds, and error bars show $\pm$ one
    standard deviation across seeds. Dashed horizontal lines
    indicate the corresponding DPO and PPO baseline means,
    with shaded bands showing $\pm$ one standard deviation.}
    \label{fig:schedules qwen}
\end{figure*}

\begin{figure*}[t]
    \centering
    \begin{subfigure}[t]{0.49\textwidth}
        \centering
        \includegraphics[width=\textwidth]{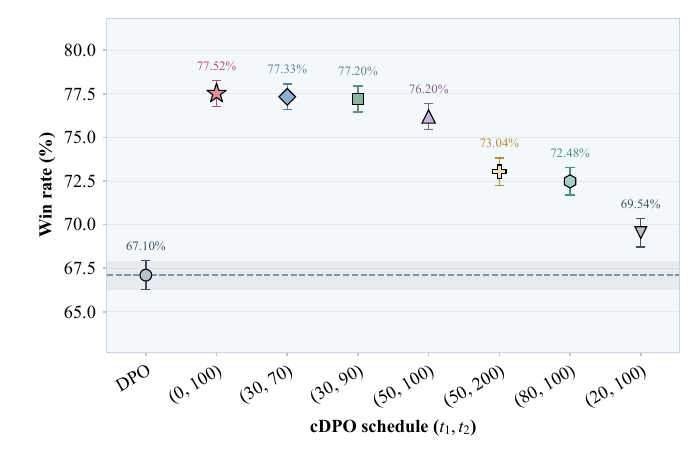}
        \caption{cDPO schedules $(t_1,t_2)$}
        \label{fig:schedules llama cdpo}
    \end{subfigure}\hfill
    \begin{subfigure}[t]{0.49\textwidth}
        \centering
        \includegraphics[width=\textwidth]{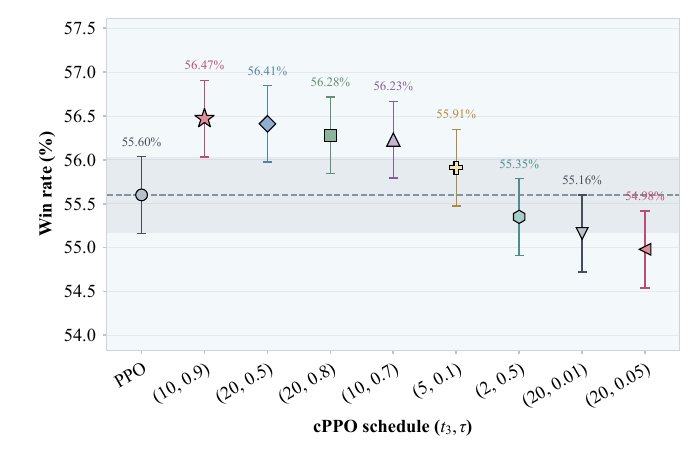}
        \caption{cPPO schedules $(t_3,\tau)$}
        \label{fig:schedules llama cppo}
    \end{subfigure}
    \caption{\textbf{Controlled Variants on Llama3-8B.}
    We compare AlpacaEval win rates across different cDPO and cPPO
    schedules. Markers indicate mean test win rates across three
    independent training seeds, and error bars show $\pm$ one
    standard deviation across seeds. Dashed horizontal lines
    indicate the corresponding DPO and PPO baseline means,
    with shaded bands showing $\pm$ one standard deviation.}
    \label{fig:schedules llama}
\end{figure*}

For DPO and PPO, hyperparameters and checkpoints are selected by
win rate against the corresponding supervised fine-tuning baseline
on held-out UltraFeedback prompts, without using AlpacaEval test scores.
For GRPO, hyperparameters and checkpoints are selected on
validation data disjoint from both the training and test sets,
using the same metrics as in the reported evaluations:
mean programmatic reward for function calling, task-success rate
for $\tau^2$ Retail, and pass@1 for mathematical reasoning.
Figures~\ref{fig:schedules pythia}--\ref{fig:schedules llama}
report additional AlpacaEval results for different cDPO and cPPO
schedules on Pythia-2.8B, Qwen3-1.7B, and Llama3-8B.
Importantly, the improvements are not confined to the selected schedules.
Multiple alternative configurations also outperform the corresponding
baselines across all three models.
For example, alternative cDPO schedules achieve win rates of
$83.56$--$87.91\%$ on Qwen3-1.7B and $69.54$--$77.52\%$
on Llama3-8B, compared with baseline scores of $72.14\%$
and $67.10\%$, respectively. Several alternative cPPO schedules likewise improve over their baselines
across the three models.

\clearpage
\section{Mathematical Analysis of Endpoint Probing}
\label{app:t}

This appendix establishes the local approximation underlying endpoint probing and gives sufficient conditions under which sparse checkpoint scores remain informative between observations.

Throughout, we fix the endpoint dataset $\mathcal D_{\mathrm{end}}$ and its loss $\mathcal L_{\mathrm{end}}$, and compute all gradients with respect to the same trainable parameters. For brevity, we write
\[
e_t=\nabla_{\boldsymbol\theta}\mathcal L_{\mathrm{end}}(\boldsymbol\theta^{(t)}),
\qquad
g_t=\nabla_{\boldsymbol\theta}\mathcal L(\boldsymbol\theta^{(t)}),
\qquad
\delta_t=e_t^\top g_t,
\]
where $e_t$ and $g_t$ are the endpoint and objective gradients at checkpoint $t$, respectively, and $\delta_t$ is the corresponding endpoint-probing score.

\subsection{Local Error of the First-Order Approximation}
\label{app:endpoint-approximation}
We derive the first-order approximation underlying endpoint probing and its local error bound, mathematically clarifying how accurately $\eta\delta_t$ approximates the exact one-step change $\Delta_t$.
We make the standard local smoothness assumption that $\mathcal L_{\mathrm{end}}$ has a Lipschitz-continuous gradient on a neighborhood containing the update segment, i.e.,
\begin{equation}
    \|\nabla\mathcal L_{\mathrm{end}}(u)-\nabla\mathcal L_{\mathrm{end}}(v)\| \leq H_{\mathrm{end}}\|u-v\|,\label{eq:endpoint-local-smoothness}
\end{equation}
where $H_{\mathrm{end}}>0$ is the local smoothness constant and $\|\cdot\|$ denotes the Euclidean norm.
\begin{proposition}
\label{prop:endpoint-one-step}
For a gradient step $\boldsymbol\theta^{(t)}-\eta g_t$ with $\eta>0$, define the endpoint-loss reduction as
\[
\Delta_t
=
\mathcal L_{\mathrm{end}}(\boldsymbol\theta^{(t)})
-
\mathcal L_{\mathrm{end}}(\boldsymbol\theta^{(t)}-\eta g_t).
\]
Under the local smoothness assumption in Eq.~\eqref{eq:endpoint-local-smoothness}, we have
\[
|\Delta_t-\eta\delta_t|
\leq \frac{H_{\mathrm{end}}\eta^2}{2}\|g_t\|^2.
\]
Hence, $\delta_t$ correctly predicts the direction of the loss change whenever
\[
|\delta_t|>\frac{H_{\mathrm{end}}\eta}{2}\|g_t\|^2.
\]
\end{proposition}
\begin{proof}
Let $\boldsymbol\theta(s)=\boldsymbol\theta^{(t)}-s\eta g_t$ for $s\in[0,1]$. Since $\frac{\mathrm d}{\mathrm ds}\mathcal L_{\mathrm{end}}(\boldsymbol\theta(s))=-\eta\nabla\mathcal L_{\mathrm{end}}(\boldsymbol\theta(s))^\top g_t$, integrating along the update segment and subtracting $\eta\delta_t$ gives
\[
\Delta_t-\eta\delta_t=\eta\int_0^1\big[\nabla\mathcal L_{\mathrm{end}}(\boldsymbol\theta(s))-\nabla\mathcal L_{\mathrm{end}}(\boldsymbol\theta^{(t)})\big]^\top g_t\,\mathrm ds.
\]
The distance from $\boldsymbol\theta(s)$ to the initial checkpoint is $s\eta\|g_t\|$. Local smoothness therefore implies
\[
\big\|\nabla\mathcal L_{\mathrm{end}}(\boldsymbol\theta(s))-\nabla\mathcal L_{\mathrm{end}}(\boldsymbol\theta^{(t)})\big\|\leq H_{\mathrm{end}}s\eta\|g_t\|.
\]
Applying the triangle inequality and Cauchy--Schwarz, we obtain
\begin{align*}
|\Delta_t-\eta\delta_t|\leq\eta\int_0^1
\big\|\nabla\mathcal L_{\mathrm{end}}(\boldsymbol\theta(s)) -\nabla\mathcal L_{\mathrm{end}}(\boldsymbol\theta^{(t)})\big\|\|g_t\|\,\mathrm ds&\leq H_{\mathrm{end}}\eta^2\|g_t\|^2\int_0^1 s\,\mathrm ds\\
&=\frac{H_{\mathrm{end}}\eta^2}{2}\|g_t\|^2.
\end{align*}
Moreover, if $|\delta_t|>H_{\mathrm{end}}\eta\|g_t\|^2/2$, then
$|\Delta_t-\eta\delta_t|<|\eta\delta_t|$. Thus, the approximation error cannot reverse the sign of $\eta\delta_t$, and $\Delta_t$ and $\delta_t$ have the same sign.
\end{proof}

\textbf{Learning-rate dependence.}
At a fixed checkpoint and for a fixed update direction, the absolute approximation error is bounded by a term quadratic in $\eta$. If $\delta_t\neq0$, the relative error satisfies
\begin{equation}
\frac{|\Delta_t-\eta\delta_t|}{\eta|\delta_t|}\leq\frac{H_{\mathrm{end}}\eta\|g_t\|^2}{2|\delta_t|}.
\end{equation}
When the right-hand side is below $1$, the error is smaller than the magnitude of the first-order prediction, so $\Delta_t$ and $\delta_t$ have the same sign. Thus, a sufficiently small learning rate ensures sign agreement for a fixed nonzero score. Predictions near zero require more caution because even a small approximation error may reverse their sign.

\textbf{Gradient-norm clipping.}
For a clipping threshold $C>0$ for the gradient norm, define
\[
 \alpha_t=\min\Big(1,\frac{C}{\|g_t\|}\Big),
 \qquad \bar g_t=\alpha_t g_t,
\]
with $\alpha_t=1$ when $g_t=0$. Let $\Delta_t^{\mathrm{clip}}$ be the endpoint-loss
reduction after $\boldsymbol\theta^{(t)}-\eta\bar g_t$. Proposition~\ref{prop:endpoint-one-step} implies
\begin{equation}
 \big|\Delta_t^{\mathrm{clip}}-\eta\alpha_t\delta_t\big|
 \leq\frac{H_{\mathrm{end}}\eta^2}{2}
       \min(\|g_t\|^2,C^2).
 \label{eq:clipped-endpoint-error}
\end{equation}
Gradient clipping scales the update by $\alpha_t$, so the first-order prediction becomes $\eta\alpha_t\delta_t$ without changing its sign. It also reduces the approximation-error bound by a factor of $\alpha_t^2$, acting as a smaller effective learning rate~\citep{zhang2019gradient}.

\textbf{Optimizer-aware interpretation.}
Endpoint probing isolates the gradient contribution of each objective component at a given checkpoint, regardless of the optimizer used during training. Its additive decomposition therefore remains valid for models trained with AdamW. Specifically, $\delta_t$ measures whether the current objective gradient supports the endpoints, as formalized by the gradient step in Eq.~\eqref{eq:endpoint-change}.
The same analysis also accommodates optimizer-specific updates. For an actual update $\boldsymbol\theta^{(t+1)}=\boldsymbol\theta^{(t)}-\eta_t d_t$, local smoothness gives
\begin{equation}
\big|
\mathcal L_{\mathrm{end}}(\boldsymbol\theta^{(t)})
-\mathcal L_{\mathrm{end}}(\boldsymbol\theta^{(t+1)})
-\eta_t e_t^\top d_t
\big|
\leq
\frac{H_{\mathrm{end}}\eta_t^2}{2}\|d_t\|^2.
\label{eq:actual-update-bound}
\end{equation}
For AdamW, $d_t$ incorporates momentum, adaptive scaling, and weight decay, and may therefore differ from $g_t$. Our analysis focuses on the objective gradients to separate positive and negative learning, while predicting an actual optimizer step requires using $d_t$ instead.
A smaller learning rate and a smaller update-direction norm $\|d_t\|$ likewise tighten the approximation-error bound in this setting.

\subsection{Impact of Sparse Checkpoint Sampling on Probing}
\label{app:theory 1}

We establish sufficient conditions under which a checkpoint score represents the direction of endpoint alignment between observations. With the endpoint dataset fixed, we consider an interval over which the probed objective is also fixed and has a locally Lipschitz-continuous gradient:
\begin{equation}
\|g_{k+1}-g_k\|
\leq
H_g\|\boldsymbol\theta^{(k+1)}-\boldsymbol\theta^{(k)}\|,
\label{eq:objective-gradient-smoothness}
\end{equation}
where $H_g\geq0$ is a local smoothness constant. This assumption controls gradient changes caused by parameter movement. It does not automatically hold across changes in rollout data or advantages in PPO and GRPO; applying the result across such changes requires the same bound to remain valid.

\begin{proposition}
\label{prop:endpoint-persistence}
Suppose that Eqs.~\eqref{eq:endpoint-local-smoothness}
and~\eqref{eq:objective-gradient-smoothness} hold throughout the interval from $t$ to $s$, with
$\|e_k\|\leq B_{\mathrm{end}}$ and $\|g_k\|\leq B_g$.
Then
\begin{equation}
|\delta_s-\delta_t|
\leq
\bigl(H_{\mathrm{end}}B_g+B_{\mathrm{end}}H_g\bigr)
\sum_{k=t}^{s-1}
\|\boldsymbol\theta^{(k+1)}-\boldsymbol\theta^{(k)}\|.
\label{eq:score-persistence}
\end{equation}
If the right-hand side is smaller than $|\delta_t|$, the score retains the sign of $\delta_t$ at every step from $t$ through $s$.
\end{proposition}

\begin{proof}
Using $\delta_k=e_k^\top g_k$, we write
\[
\delta_{k+1}-\delta_k
=
(e_{k+1}-e_k)^\top g_{k+1}
+
e_k^\top(g_{k+1}-g_k).
\]
The triangle inequality and Cauchy--Schwarz give
\[
|\delta_{k+1}-\delta_k|
\leq
\|e_{k+1}-e_k\|\|g_{k+1}\|
+
\|e_k\|\|g_{k+1}-g_k\|.
\]
Endpoint-loss smoothness implies
\[
\|e_{k+1}-e_k\|
\leq
H_{\mathrm{end}}
\|\boldsymbol\theta^{(k+1)}-\boldsymbol\theta^{(k)}\|.
\]
Combining this with the objective-gradient assumption and the norm bounds yields
\[
|\delta_{k+1}-\delta_k|
\leq
\bigl(H_{\mathrm{end}}B_g+B_{\mathrm{end}}H_g\bigr)
\|\boldsymbol\theta^{(k+1)}-\boldsymbol\theta^{(k)}\|.
\]
Summing over the interval, we obtain
\[
|\delta_s-\delta_t|
=
\big|\sum_{k=t}^{s-1}(\delta_{k+1}-\delta_k)\big|
\leq
\sum_{k=t}^{s-1}|\delta_{k+1}-\delta_k|
\leq
\bigl(H_{\mathrm{end}}B_g+B_{\mathrm{end}}H_g\bigr)
\sum_{k=t}^{s-1}
\|\boldsymbol\theta^{(k+1)}-\boldsymbol\theta^{(k)}\|.
\]
For any intermediate step $j\in\{t,\ldots,s\}$, the same argument bounds
$|\delta_j-\delta_t|$ by the corresponding sum up to $j-1$.
Because all summands are nonnegative, this partial sum cannot exceed the bound for the full interval.
If the full bound is smaller than $|\delta_t|$, then
$|\delta_j-\delta_t|<|\delta_t|$, so $\delta_j$ cannot reach or cross zero.
\end{proof}

\textbf{Learning-rate dependence.}
When the trajectory follows gradient descent on the probed objective,
$\boldsymbol\theta^{(k+1)}=\boldsymbol\theta^{(k)}-\eta_k g_k$.
Using $\|g_k\|\leq B_g$, we obtain
\begin{equation}
|\delta_s-\delta_t|
\leq
\bigl(H_{\mathrm{end}}B_g+B_{\mathrm{end}}H_g\bigr)
B_g\sum_{k=t}^{s-1}\eta_k.
\end{equation}
For a constant learning rate $\eta$, the bound is proportional to $(s-t)\eta$.
With the smoothness and norm bounds held fixed, smaller learning rates and shorter checkpoint intervals tighten this upper bound.

\textbf{Gradient-norm clipping.}
For gradient descent on the probed objective with clipping threshold $C>0$,
\[
\|\boldsymbol\theta^{(k+1)}-\boldsymbol\theta^{(k)}\|
=
\eta_k\min(\|g_k\|,C)
\leq
\eta_k\min(B_g,C).
\]
Consequently,
\begin{equation}
|\delta_s-\delta_t|
\leq
\bigl(H_{\mathrm{end}}B_g+B_{\mathrm{end}}H_g\bigr)
\min(B_g,C)\sum_{k=t}^{s-1}\eta_k.
\end{equation}
Clipping tightens this upper bound when $C<B_g$. If the bound is below $|\delta_t|$, the score retains its sign throughout the interval.
For AdamW, or when probing an individual component along a jointly trained trajectory, Eq.~\eqref{eq:score-persistence} instead uses the actual parameter displacements.
These results provide sufficient conditions for interpreting sparse scores as interval-level trends; they do not establish that the checkpoint intervals used in our experiments satisfy these conditions.

\section{Limitations and Further Discussion}\label{sec:limitations}

Endpoint probing examines whether locally induced target shaping persists as a stable supervisory signal, using final-model responses as empirical surrogates for implicit targets. These responses need not be optimal to provide a common reference for comparing components across checkpoints. Our findings characterize stable and phase-dependent target shaping relative to the evaluated in-distribution and out-of-distribution endpoints.
Endpoint alignment alone does not determine task performance. In DPO, removing early negative learning or late positive learning reduced win rates despite their opposition to endpoint fitting, supporting a beneficial regularization-like role consistent with related work~\citep{han2020sigua}. In PPO and GRPO, negative learning counteracts endpoint-aligned positive learning, consistent with an exploration-related role and prior uses of response suppression to encourage alternative generations~\citep{chen2026eepo}.
To the best of our knowledge, these interpretations offer the most plausible account of the observed dynamics given the available evidence, although they may be revised as the understanding of negative learning develops. The controlled-method results demonstrate the practical value of adjusting component strengths within the evaluated settings, without requiring these interpretations to be definitive.

\clearpage

\end{document}